\documentclass{article}

\usepackage[preprint]{neurips_2024}

\usepackage[utf8]{inputenc} 
\usepackage[T1]{fontenc}    
\usepackage{hyperref}       
\usepackage{url}            
\usepackage{booktabs}       
\usepackage{amsfonts}       
\usepackage{nicefrac}       
\usepackage{microtype}      
\usepackage{xcolor}         
\usepackage{graphicx}
\usepackage{subcaption} 
\usepackage{amsmath}
\newtheorem{proposition}{Proposition}

\title{Mind the Gap: Continuous Magnification Sampling for Pathology Foundation Models}

\author{%
  \textbf{Alexander Möllers}$^{1,2,3}$\thanks{Corresponding author. Email: \url{a.moellers@tu-berlin.de}. Code, models and new benchmarks (TCGA-MS, BRACS-MS) to reproduce the figures in this paper at: \url{https://github.com/bifold-pathomics/continuous-magnification-sampling}.} \quad
  \textbf{Julius Hense}$^{1,2}$ \quad
  \textbf{Florian Schulz}$^{1,2}$ \\[0.5em]
  \textbf{Timo Milbich}$^{3}$ \quad
  \textbf{Maximilian Alber}$^{3,4}$ \quad
  \textbf{Lukas Ruff}$^{3}$ \\[1.2em]
  {\small $^1$Berlin Institute for the Foundations of Learning and Data (BIFOLD)} \\[0.2em]
  {\small $^2$Machine Learning Group, Technische Universität Berlin} \\[0.2em]
  {\small $^3$Aignostics} \\[0.2em]
  {\small $^4$Charité, Universitätsmedizin Berlin}
}

\begin{document}

\maketitle

\begin{abstract}
In histopathology, pathologists examine both tissue architecture at low magnification and fine-grained morphology at high magnification. Yet, the performance of pathology foundation models across magnifications and the effect of magnification sampling during training remain poorly understood. We model magnification sampling as a multi-source domain adaptation problem and develop a simple theoretical framework that reveals systematic trade-offs between sampling strategies. We show that the widely used discrete uniform sampling of magnifications (0.25, 0.5, 1.0, 2.0 mpp) leads to degradation at intermediate magnifications. We introduce continuous magnification sampling, which removes gaps in magnification coverage while preserving performance at standard scales. Further, we derive sampling distributions that optimize representation quality across magnification scales. To evaluate these strategies, we introduce two new benchmarks (TCGA-MS, BRACS-MS) with appropriate metrics. Our experiments show that continuous sampling substantially improves over discrete sampling at intermediate magnifications, with gains of up to 4 percentage points in balanced classification accuracy, and that optimized distributions can further improve performance. Finally, we evaluate current histopathology foundation models, finding that magnification is a primary driver of performance variation across models. Our work paves the way towards future pathology foundation models that perform reliably across magnifications.
\end{abstract}

\section{Introduction}
\label{sec1}

Foundation models are seeing widespread adoption in digital pathology and are increasingly embedded in diagnostic assistants, interactive AI tools, and digital slide viewers that operate across the continuous magnification spectrum \citep{pathchat, albastaki2025multi}. However, how magnification sampling during pretraining affects representation quality remains underexplored. 

In practice, models are trained on a discrete set of scanner resolutions and image patches are usually sampled uniformly from the standard magnifications of 0.25, 0.5, 1.0 and 2.0 microns per pixel (mpp). Although this approach is widely adopted and has been reported to improve performance across benchmarks \citep[e.g.,][]{ciga_self_2022,kang_benchmarking_2023,ai_towards_2024, karasikov2025trainingstateoftheartpathologyfoundation},  it lacks a principled justification.

In this paper, we model magnification sampling as a multi-source domain adaptation problem and develop a theoretical framework that reveals trade-offs between different sampling strategies. Importantly, we find that discrete uniform sampling fails to ensure uniform representation quality across different scales, and optimizes neither the average nor the worst-case quality across magnifications. Our analysis predicts specific failure modes, including representation degradation at intermediate and boundary magnifications.  We validate these predictions empirically and find that they manifest both in controlled experiments and in state-of-the-art foundation models and find that model rankings on benchmarks can strongly depend on the image magnification.

\begin{figure*}[t!]
\centering
\includegraphics[width=\textwidth]{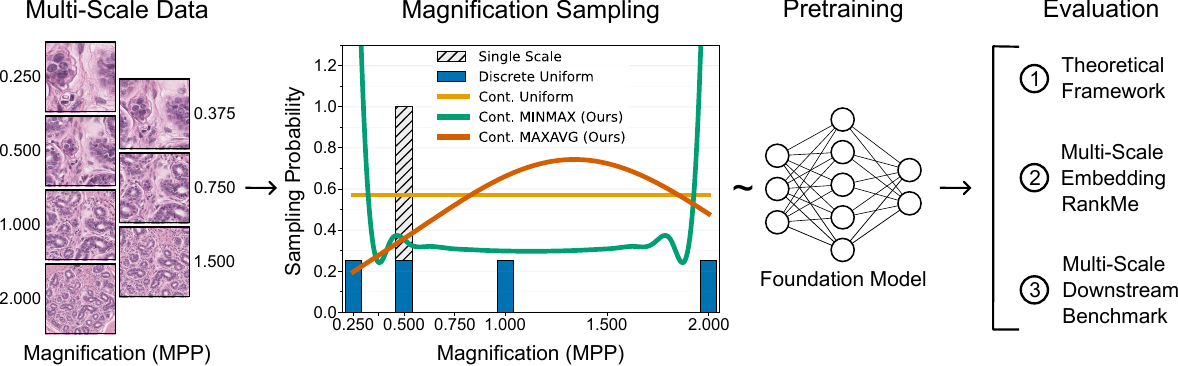}
\caption{Overview of our approach. We derive novel, principled magnification sampling strategies to pretrain vision foundation models (FMs) from multi-scale histopathology data and compare them to established discrete magnification sampling protocols (MPP = \textit{microns per pixel}). For evaluation, we introduce three techniques to measure FM embedding quality across magnifications: (1)~a data-free theoretical domain adaptation framework for magnification sampling, (2)~a RankMe score to quantify the information richness of a model’s embedding space, and (3)~a multi-scale predictive performance benchmark for FM-derived downstream models. Together, these evaluations provide a comprehensive view of the magnification performance profile of new and established pathology foundation models.}
\label{fig:explantatory_figure}
\end{figure*}

Building on this analysis, we propose continuous magnification sampling. Rather than restricting training to fixed scanner resolutions, we synthesize patches at arbitrary magnifications via dynamic crop-and-resize operations during training. This requires no architectural changes, adds no computational overhead, and integrates seamlessly with existing self-supervised frameworks like DINOv2 \citep{oquab2024dinov2learningrobustvisual}. Perhaps surprisingly, continuous sampling does not sacrifice performance at standard scales while improving performance by up to 4 percentage points in balanced classification accuracy at intermediate magnifications in our controlled experiments. We further derive principled non-uniform distributions that yield additional gains on downstream benchmarks.
To evaluate the impact of different sampling strategies on a model's embedding space across magnifications we use the unsupervised RankMe metric \citep{Garrido_2023}, a strong indicator for downstream task performance \citep[e.g.,][]{pmlr-v224-ericsson23a, jaume2024hest1kdatasetspatialtranscriptomics, ai_towards_2024}. To estimate the effect of different strategies on clinical downstream tasks we release a set of multi-scale benchmarks (TCGA-MS, BRACS-MS). Figure~\ref{fig:explantatory_figure} provides an overview of our approach. We summarize our contributions as follows:

\begin{itemize}
\item \textbf{Theoretical framework}: We model magnification sampling as a multi-source domain adaptation problem and show that discrete uniform sampling, the prevalent strategy in the literature, suffers from systematic deficiencies including degradation at intermediate and boundary magnifications.
\item \textbf{Continuous magnification sampling}: We propose continuous magnification training, which improves performance at intermediate magnifications without sacrificing performance at standard scales. We derive non-uniform sampling distributions that yield further gains.
\item \textbf{Evaluation methodology and benchmarks}: We introduce a RankMe-based profiling approach and release two new multi-scale benchmarks (TCGA-MS, BRACS-MS) for downstream evaluation. Analyzing current state-of-the-art foundation models, we find substantial variation in their performance profiles across magnifications and degradation patterns that are consistent with our theoretical predictions.
\end{itemize}

\section{Magnification Sampling as Domain Adaptation} \label{sec:theoretical_framework}

Current pathology foundation models are trained on patches sampled from multiple discrete magnifications, yet the impact of this sampling strategy on learned representations remains poorly understood. In this section, we develop a theoretical framework that models magnification sampling as a multi-source domain adaptation problem. We use this perspective to systematically analyze different sampling strategies and to propose principled continuous alternatives.

\subsection{A Domain Adaptation Framework for Magnification Sampling}

Domain adaptation studies how representations learned on source domains transfer to target domains. In our setting, each magnification constitutes a source domain, and we define our target as strong representation quality across the entire continuous magnification range (0.25–2.0 mpp). To model how training signal accumulates across magnifications, we make two assumptions:

\begin{itemize}
     \item \textbf{Coverage-Based Proxy:} We expect the representation quality at magnification $y$ to improve with the number of training samples seen at nearby magnifications. 
    \item \textbf{Domain Transfer Assumption:} We assume that representations learned at one magnification transfer to nearby magnifications with decreasing effectiveness as the magnification distance increases. We model this via a similarity kernel $K(x,y)$, where $x$, $y$ are the mpp values of the patches. 
\end{itemize}

Combining these assumptions, we model the accumulated training signal $S(y)$, which captures the training signal that reaches magnification $y$ after accounting for transfer from nearby scales, as:
\begin{equation}\label{eq:domain_adaption}
    S(y) =  \int_{0.25}^{2} p(x) K(x,y) \, dx,
\end{equation}
where $p(x)$ is the sampling distribution that assigns a probability to each magnification.  Intuitively, $S(y)$ is high when we sample many patches close to $y$ and low when we sample none in its vicinity. Since representation quality improves with the training signal, $S(y)$ can be considered a proxy for model performance at different magnifications. We analyze the model for two different Kernels that encode different notions of similarity in pathology:

\paragraph{Absolute Distance Kernel}
The absolute distance kernel $K_{\text{abs}}(x,y) = \frac{1}{1 + |x-y|}$ encodes that the effect of sampling a patch at magnification $x$ on the representation at magnification $y$ decreases proportionally to the magnification distance $|x-y|$.
\paragraph{Information-Based Kernel} Defined as $K_{\text{info}}(x,y) = \left(\frac{\min(x,y)}{\max(x,y)}\right)^2$, it models information transfer through field of view overlap.  A 224×224 patch at magnification $x$ covers a tissue area of $(224*x)^2$ square microns. The kernel represents the fraction of overlapping field of view when the same tissue is viewed at magnification $y$.

In Section~\ref{sec:experiments}, we empirically validate the modeling assumptions made in this section by investigating the embedding space of trained models, showing how similar magnifications are close while more distant ones are further apart (Figure~\ref{fig:rep_quality_combined}).

\subsection{Transfer Potential and Magnification Prototypes} \label{sec:transfer_potential}

By framing magnification sampling as a domain adaptation problem, we can reason about properties of different magnification sampling distributions. A natural question that arises is whether samples on all magnifications are equally valuable. We argue that they are not. Samples at central magnifications share visual characteristics with many other scales and are the most prototypical. In contrast, samples at boundary magnifications lie at the extremes of the range, and their training signal transfers only to neighboring magnifications in one direction.

\begin{figure*}[t!]
\centering
\begin{subfigure}[b]{0.48\textwidth}
    \centering
    \includegraphics[width=\textwidth]{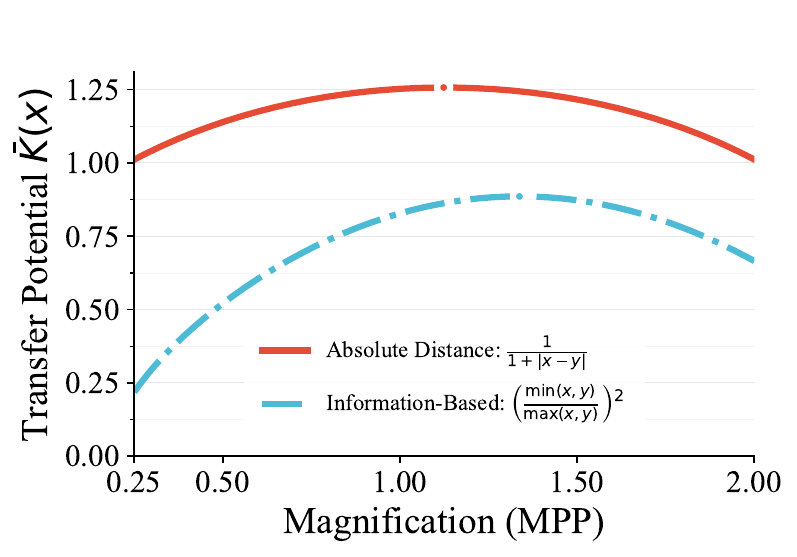}
    \caption{}
    \label{fig:transfer_potential_kernel}
\end{subfigure}
\hfill
\begin{subfigure}[b]{0.48\textwidth}
    \centering
    \includegraphics[width=\textwidth]{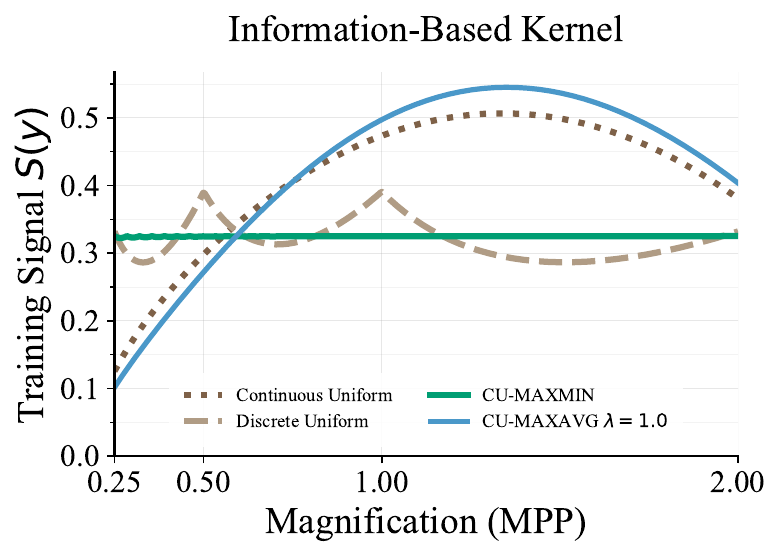}
    \caption{}
    \label{fig:rep_quality}
\end{subfigure}
\caption{Properties of the domain adaptation framework. (a) Transfer potential $\bar{K}(x)$ for three similarity kernels across the magnification range. Central magnifications exhibit higher transfer potential due to shared visual features with neighboring scales, while boundary magnifications show reduced potential. (b) Accumulated Training Signal $S(y)$ for the ``Information-Based Kernel'' for the continuous uniform and discrete sampling distributions. Continuous uniform sampling leads to systematic degradation at boundaries (Proposition \ref{prop:mag_prototypes}), while discrete sampling creates gaps at intermediate magnifications.}
\label{fig:sampling_theory}
\end{figure*}

To see this formally, we can examine how much training signal is accumulated across all target magnifications:
\begin{align}
S(p) &= \int_{0.25}^{2} \int_{0.25}^{2} p(x) K(x,y) \, dx \, dy \\
&= \int_{0.25}^{2} p(x)\int_{0.25}^{2} K(x,y) \, dy \, dx \\
&= \int_{0.25}^{2} p(x) \bar{K}(x) \, dx,
\end{align}
where the transfer potential $\bar{K}(x) = \int_{0.25}^{2}  K(x,y) \, dy$ quantifies how much the signal from sampling a data point at one magnification $x$ improves the representation quality at all other magnifications. Figure~\ref{fig:transfer_potential_kernel} illustrates the transfer potential for several kernel choices: all peak at central magnifications and decay towards the boundaries. We can formalize this property for large classes of kernels and provide the proof in \ref{apd:domain_adaption_framework}:

\begin{proposition} [Magnification Prototypes]
For any symmetric kernel $K(x,y)=f(|x-y|)$ with $f$
monotonically decreasing, the transfer potential of a data point $\bar{K}(x) = \int_{b}^{a}  K(x,y) \, dy$ is maximized at $x= \frac{a+b}{2}$ and minimized at the boundaries $x \in \{a,b\}$.
\label{prop:mag_prototypes}
\end{proposition}

\subsection{Continuous Sampling}\label{sec:cms}

Our framework, with its mild assumptions and basic properties, enables us to analyze the training signal that reaches all magnifications under the discrete uniform magnification sampling strategy commonly used in pathology foundation model pre-training. We can immediately see that for discrete sampling less training signal reaches the intermediate magnifications not present in training (Figure~\ref{fig:rep_quality}).

To address this problem, we introduce \textit{continuous magnification training} through dynamic patch generation. By extracting larger source patches from existing WSIs and applying controlled crop-and-resize operations, we can synthesize training patches at arbitrary target magnifications. Formally, to create a patch at target mpp $t$ from a source patch at mpp $s$, we crop a region of size:
$$cs_{source} = cs_{target} \times \frac{t}{s}$$
and resize it to the desired patch size $cs_{target}$.  During training, for each incoming patch at a standard magnification, we sample the target magnification from a continuous distribution:
$$t \sim p(\text{mpp})$$

The continuous magnification sampling strategy resolves the issue of representation quality collapsing at intermediate magnifications (Figure~\ref{fig:rep_quality}), but it does not account for the varying transfer potential across scales (see Section~\ref{sec:transfer_potential}). This begs the question whether sampling can be improved even further.

\subsection{Optimizing Continuous Sampling}\label{sec:uni_opt} 

We investigate what distributions emerge when we maximize the average or the minimal training signal that is accumulated across magnifications.

\begin{figure*}[t!]
\centering
\begin{subfigure}[b]{0.48\textwidth}
    \centering
    \includegraphics[width=\textwidth]{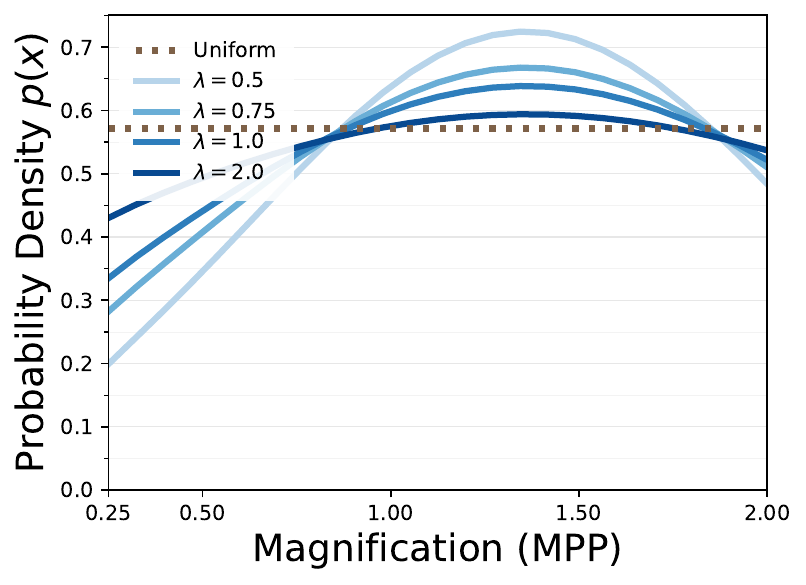}
    \caption{}
    \label{fig:max_avg_dist}
\end{subfigure}
\hfill
\begin{subfigure}[b]{0.48\textwidth}
    \centering
    \includegraphics[width=\textwidth]{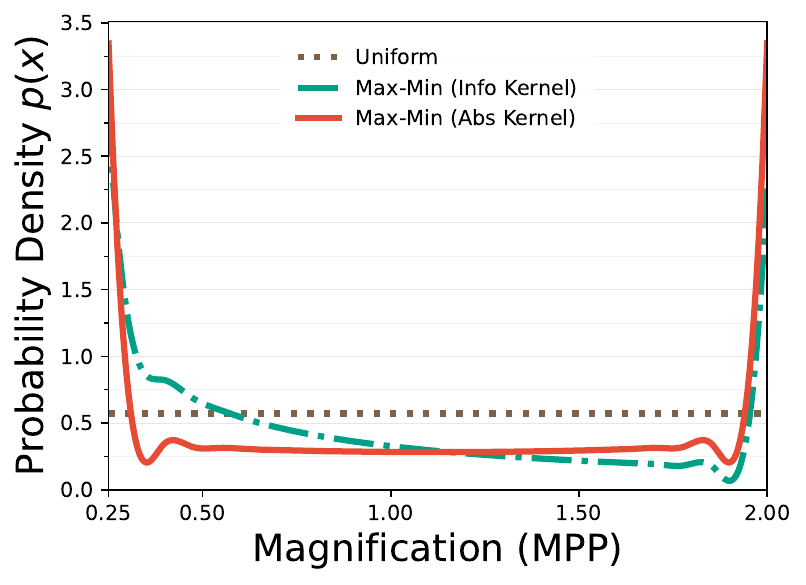}
    \caption{}
    \label{fig:max_min_dist}
\end{subfigure}
\caption{ Optimized magnification sampling distributions. (a)~Sampling distributions from max-average optimization with entropy regularization for the information-based kernel. As the regularization parameter $\lambda$ increases, the optimal distribution shifts from sampling more prototypical magnifications near the center to near-uniform. (b) Max-min optimization oversamples boundary magnifications for both kernels.}
\label{fig:optimal_distributions}
\end{figure*}

\paragraph{Maximizing Overall Representation Quality}
We search for a magnification sampling distribution $p$ that optimizes the overall accumulated training signal $S(p)$ over our target domain (i.e., the full range of magnification levels $[0.25, 2]$). We find that without further modification, this optimization trivially collapses to distributions that place all probability mass on magnifications with high transfer potential (i.e., the middle of the spectrum). To prevent this, we introduce an entropy regularization term $H[p]$ that encourages more dispersed distributions:
\begin{equation}\label{def:max_avg}
p^* = \arg\max_{p} S(p) + \lambda H[p]
\end{equation}
where the entropy is defined as:
\begin{equation}
H[p] = -\int_{0.25}^{2} p(x) \log p(x) \, dx
\end{equation}
and $\lambda > 0$ controls the trade-off between sampling high-potential magnifications and maintaining diversity. Higher values of $\lambda$ lead to more uniform magnification sampling distributions (Figure~\ref{fig:max_avg_dist}).

\paragraph{Maximizing Worst-Case Representation Quality}
An alternative objective is to maximize the minimum accumulated training signal across magnifications:
\begin{equation}
p^* = \arg\max_{p} \min_{y \in [0.25, 2]} S(y)
\label{def:max_min}
\end{equation} 
This formulation seeks distributions that avoid failure modes at any particular scale. We discretize the problem and solve it using standard optimization techniques. The resulting distributions consistently oversample boundary magnifications (Figure~\ref{fig:max_min_dist}). 

Plotting the accumulated training signal for both the max-average (CU-MAXAVG) and max-min (CU-MINMAX) distributions in Figure~\ref{fig:sampling_theory}, we find that the max-min solution accumulates training signal nearly uniformly across magnifications. In contrast, the max-average solution concentrates signal at central magnifications while accepting weaker coverage at the boundaries. Nevertheless, it leads to the highest amount of accumulated signal on average. This illustrates a natural trade-off between worst-case and average-case optimization. In Section~\ref{sec:experiments}, we investigate this trade-off empirically and find that it manifests in the embedding space as a form of dimensional collapse. Models trained with CU-MAXAVG achieve higher average embedding rank but suffer from reduced rank at boundary magnifications, while CU-MINMAX maintains more uniform embedding quality across all scales (Figure~\ref{fig:rep_quality_combined}).

\section{RankMe Profiling and Multi-Scale Benchmarks}\label{sec:ds_ms}

Evaluating foundation models across the magnification spectrum presents two challenges.
First, existing benchmarks typically provide labels only at a single magnification, most commonly 0.25 or 0.5 mpp \citep[e.g.,][]{kaiko.ai2024eva,zhang2025standardizing}. Second, most clinical tasks are inherently magnification dependent, making it difficult to disentangle embedding quality from task-specific effects. We address these challenges with two complementary contributions: (1) we propose label-free profiling of representation quality across magnifications via the RankMe metric, and (2) we introduce two new multi-scale benchmarks that span both standard and intermediate magnifications for supervised downstream model assessment.

\subsection{RankMe for Profiling Representation Quality}\label{sec:rankme}

We use the RankMe metric \citep{Garrido_2023} as a  task-agnostic profiling tool. RankMe quantifies the effective rank of a model's embedding space. Higher values indicate representations that span a larger subspace and encode richer information, while lower values suggest dimensional collapse. We choose RankMe because it has been shown to correlate strongly with downstream task performance across multiple digital pathology and domain-adaptation benchmarks \citep{pmlr-v224-ericsson23a, jaume2024hest1kdatasetspatialtranscriptomics, ai_towards_2024}.

Given a set of N patches $\mathbf{X}_{\text{mpp}} = \{x_1, ..., x_N\}$ extracted at a specific mpp, we obtain their embeddings $\mathbf{Z}_{\text{m},\text{mpp}} = [z_1, ..., z_N]^T \in \mathbb{R}^{N \times K}$ by passing them through model m, where K is the embedding dimension. Then RankMe is defined as: 

\begin{equation}\label{eqn:rankme}
\text{RankMe}(\mathbf{Z}_{\text{m},\text{mpp}}) = \exp \left( - \sum_{k=1}^{\min(N,K)} p_k \log p_k \right),
\end{equation}
where
\begin{equation}
p_k = \frac{\sigma_k(\mathbf{Z}_{\text{m},\text{mpp}})}{|\sigma(\mathbf{Z}_{\text{m},\text{mpp}})|_1} + \epsilon,
\end{equation}

where $\sigma_k(\mathbf{Z}_{\text{m},\text{mpp}})$ denotes the $k$-th singular value of the embedding matrix $\mathbf{Z}_{\text{m},\text{mpp}}$, $|\sigma(\mathbf{Z}_{m,\text{mpp}})|_1 = \sum_{i=1}^{\min(N,K)} \sigma_i(\mathbf{Z}_{m,\text{mpp}})$ is the sum of all singular values, and $\epsilon$ is a small constant for numerical stability. By evaluating models at both standard (0.25, 0.5, 1.0, 2.0 mpp) and intermediate (0.375, 0.75, 1.5 mpp) magnifications, we can profile how representation quality varies across scales.

\subsection{Multi-Scale Downstream Benchmarks}\label{sec:ms_benchmarks}
While RankMe profiles provide a task-agnostic assessment, embedding quality needs to influence downstream clinical tasks to be of practical relevance. We introduce two benchmarks designed to complement RankMe profiling by testing classification performance across the magnification spectrum. Table~\ref{tab:benchmark_stats} summarizes their characteristics.

\begin{table}[t]
\centering
\caption{Statistics of TCGA-MS and BRACS-MS, two new multi-scale benchmarks introduced in this work. Both datasets enable evaluation across seven magnifications (0.25, 0.375, 0.5, 0.75, 1.0, 1.5, 2.0 mpp), filling a gap in the current benchmark landscape. }
\label{tab:benchmark_stats}
\resizebox{\textwidth}{!}{%
\begin{tabular}{@{}lcccccc@{}}
\toprule
\textbf{Benchmark} & \textbf{Task} & \textbf{WSIs} & \textbf{Total Patches} & \textbf{Magnifications} & \textbf{Classes} & \textbf{Validation} \\
\midrule
\textbf{TCGA-MS} & Multi-Cancer & 128 & 67,200 & 7 & 16 & 5-fold CV \\
 & morphological & &&  (0.25--2.0) & & \\
 & subtyping  & & & & \\
\midrule
\textbf{BRACS-MS} & Subtyping & 269 & 10,773 & 7 & 6 & 5-fold CV \\
 & atypical lesions & & & (0.25--2.0) & & (3 seeds) \\
 & in breast cancer & & & & & \\
\bottomrule
\end{tabular}%
}
\vspace{0.5em}
\end{table}

\subsubsection{TCGA-MS: Multi-Cancer Morphological Subtyping}\label{sec:TCGA-SCALE}

We curated TCGA-MS in collaboration with a board-certified pathologist to provide a challenging classification task spanning diverse tissue morphologies, including clinically rare subtypes. The benchmark comprises 16 morphological subtypes from four TCGA tissue orgins (kidney, stomach, breast, and soft tissue), annotated according to the ICD-O-3 standard. For each morphological subtype, we selected eight whole-slide images and extracted patches within tissue boundaries at seven magnifications (0.25, 0.375, 0.5, 0.75, 1.0, 1.5, 2.0 mpp). Full subtype definitions are provided in~\ref{apd:benchmark_details}. TCGA-MS will be publicly released upon publication.

 \subsubsection{BRACS-MS: Atypical Lesion Classification}\label{sec:bracs_ms}

 BRACS-MS derives from the BReAst Carcinoma Subtyping (BRACS) dataset \citep{brancati2022bracs}, which contains images representing atypical lesions. The images in the dataset are characterized as Pathological Benign (PB), Usual Ductal Hyperplasia (UDH), Flat Epithelial Atypia (FEA), Atypical Ductal Hyperplasia (ADH), Ductal Carcinoma in Situ (DCIS) and Invasive Carcinoma (IC). The WSIs used to create BRACS are scanned at a 40x resolution and the annotated ROIs often exceed 4,000x4,000 pixels. To create BRACS-MS we selected all ROIs with a minimum size of 1,792$\times$1,792 pixels. We then extracted concentric crops of increasing size and resized them to obtain a benchmark consisting of 224x244 patches at seven magnifications from 269 WSIs. The original BRACS dataset is publicly available at \url{https://www.bracs.icar.cnr.it/}. We provide code to generate BRACS-MS from the source data in our repository.

\section{Experiments}\label{sec:experiments}

\subsection{Experimental Setup}

To investigate the impact of different magnification sampling strategies, we construct patch datasets for training at the four standard magnifications (0.25, 0.5, 1.0, and 2.0 mpp) from 200,000 whole-slide images (WSIs) from The Cancer Genome Atlas (TCGA) and Charité - Universitätsmedizin Berlin. We train ViT-S/16 models using the DinoV2 framework \citep{oquab2024dinov2learningrobustvisual} for 60,000 iterations with a batch size of 320 and report mean results with standard errors over three seeds. We use standard hyperparameters for all runs and provide full training details in \ref{apd:training_details}.
We train single-scale models that are trained exclusively on one magnification (0.25, 0.5, 1.0 and 2.0), discrete uniform models (DU) that are trained uniformly on the standard magnifications, continuous uniform models (CU) and models that optimize for the overall or the worst-case performance (Section~\ref{sec:uni_opt}). For the latter, we train models using the max-min objective (CU-MINMAX) and using the max-average objective (CU-MAXAVG). For CU-MAXAVG, we evaluated $\lambda \in \{0.5, 1.0\}$
and report results for $\lambda=1.0$, which performed marginally better on our benchmarks. Both optimized distributions are derived using the information-based kernel, which we found to more accurately capture the patterns of dimensional collapse observed in our experiments. We provide a comparison with the absolute distance kernel and further discussion of kernel choice in \ref{apd:domain_adaption_framework}.

\begin{figure}[t!]
\centering
\includegraphics[width=0.6\textwidth]{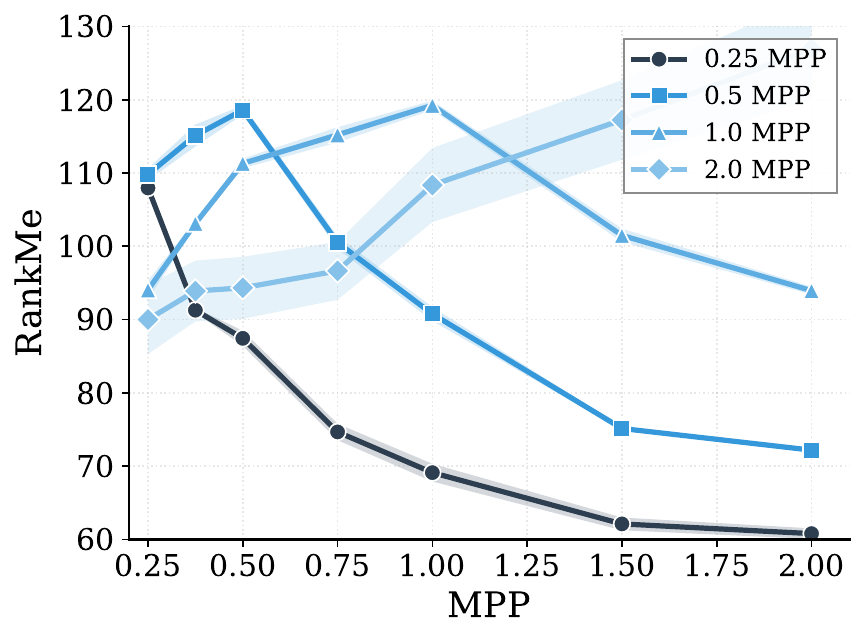}
\caption{Representation quality across magnifications for single-scale models trained on different magnifications. Each line represents a model trained at the magnification indicated in the legend. Models exhibit smooth degradation in RankMe scores as evaluation magnification moves away from training magnification.}
\label{fig:vits_rankme_ss}
\end{figure}

\subsection{Profiling Representation Quality Across Magnifications}

To investigate if and how the embedding space of different models changes depending on the sampling strategy, we extract 10,000 patches at each magnification from held-out slides, including intermediate magnifications (0.375, 0.75, 1.5 mpp). We then compute the RankMe scores for the different sampling strategies using Equation~(\ref{eqn:rankme}) from Section \ref{sec:rankme}. By plotting the mpp of the patches on the x-axis and the rank of the corresponding embeddings on the y-axis, we obtain a performance profile for each model.

Single-scale models exhibit sharp performance degradation at magnifications distant from their training scale, with rank scores dropping by up to $40\%$ (Figure~\ref{fig:vits_rankme_ss}). The lines suggest a smooth, continuous degradation in embedding quality as we move further away from the training magnification, rather than abrupt, discrete transitions.

\begin{figure*}[htbp!]
\centering
\begin{minipage}[t]{0.48\textwidth}
    \centering
    \textbf{\small A ~ RankMe per Magnification}
    \vspace{0.5mm}
    
    \includegraphics[width=\textwidth]{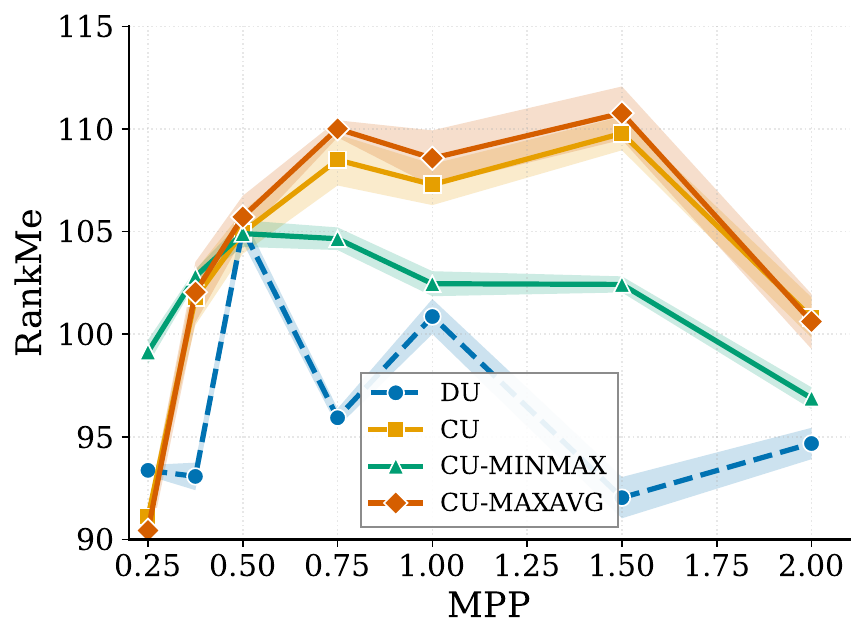}
\end{minipage}
\hfill
\begin{minipage}[t]{0.48\textwidth}
    \centering
    \textbf{\small B ~ Training Signal vs. RankME}
    \vspace{6mm}
    
    \scriptsize
    \setlength{\tabcolsep}{3pt}
    \begin{tabular}{lcccc}
    \hline
    & \multicolumn{2}{c}{$S(y)$} & \multicolumn{2}{c}{RankMe (Norm)} \\
    & Min & Tot & Min & Tot (Avg) \\
    \hline
    CU & .126 & .731 & .033 & .640 \\
    DU & .286 & .560 & .078 & .295 \\
    MAXMIN & \textbf{.322} & .570 & \textbf{.317} & .563 \\
    MAXAVG & .102 & \textbf{.755} & .000 & \textbf{.668} \\
    \hline
    \end{tabular}
\end{minipage}

\vspace{1em}

\begin{minipage}[t]{0.48\textwidth}
    \centering
    \textbf{\small C ~ Embedding Similarity (DU)}
    \vspace{0.5mm}
    
    \includegraphics[width=\textwidth]{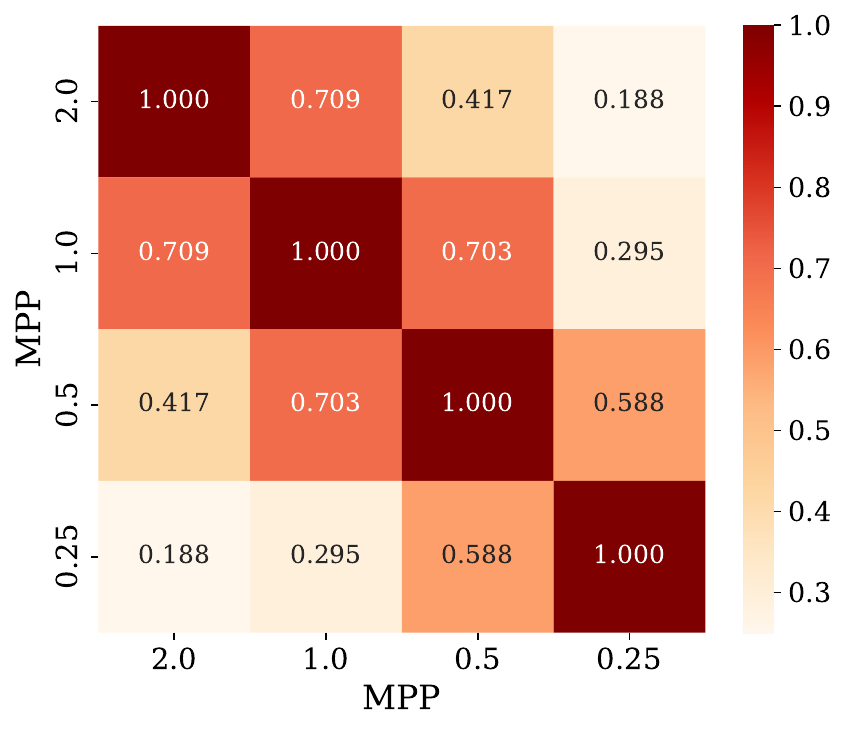}
\end{minipage}
\hfill
\begin{minipage}[t]{0.48\textwidth}
    \centering
    \textbf{\small D ~ Embedding Space (DU)}
    \vspace{0.5mm}
    
    \includegraphics[width=\textwidth]{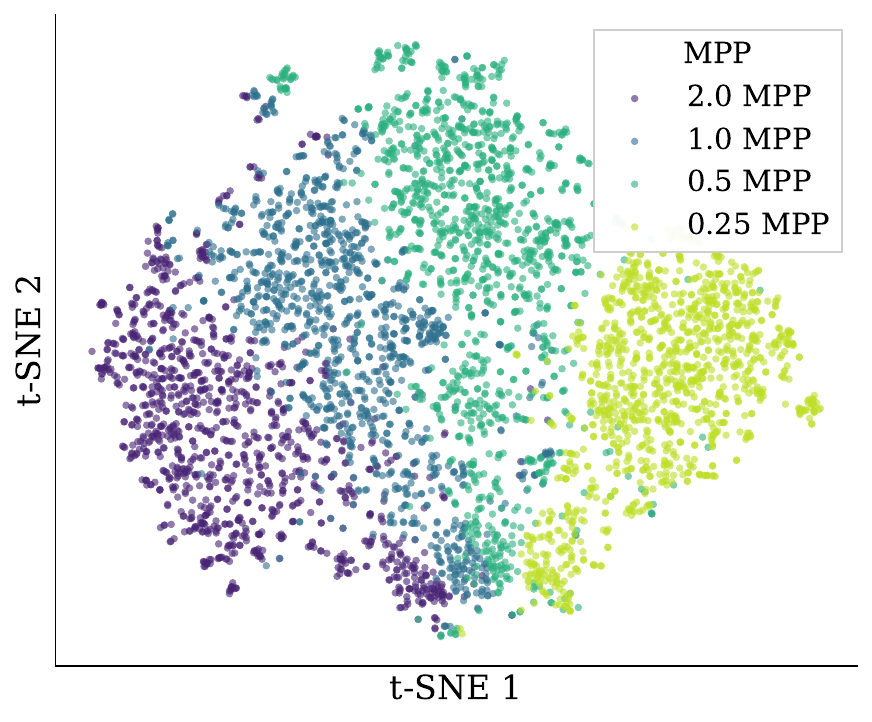}
\end{minipage}

\caption{Representation quality and embedding space organization in multi-scale models. 
(A) RankMe profiles across magnifications: discrete uniform sampling (DU) shows dips at intermediate scales, while continuous strategies produce smooth profiles. 
(B) Minimal and overall theoretical training signal and normalized RankMe scores across sampling strategies. The close alignment between theoretical predictions and measured embedding quality validates our domain adaptation framework. 
(C) Cosine similarities between embedding centroids at standard magnifications (DU model). Patches from similar magnifications cluster more closely, while larger magnification differences correspond to greater separation in the learned feature space.
(D) t-SNE projection of patch embeddings colored by magnification (DU model), showing that scale acts as a continuous organizing dimension.}
\label{fig:rep_quality_combined}
\end{figure*}

Continuing our analysis with multi-scale models (Figure~\ref{fig:rep_quality_combined}), we observe that models trained with discrete uniform (DU) magnification sampling exhibit a more robust representation quality profile than the single-scale models overall, but reveal a distinctive sawtooth pattern. Specifically, although multi-scale models avoid extreme degradation at any tested magnification, they show notable dips at intermediate scales absent from their training data. The effect is particularly pronounced at 1.5 mpp, which lies furthest from any training magnification. In contrast, models trained with continuous sampling strategies do not suffer from these failure modes at intermediate magnifications and exhibit smoother profiles of representation quality across magnifications (Figure~\ref{fig:rep_quality_combined}A). The models trained with continuous uniform sampling (CU) and the optimized variants (CU-MINMAX, CU-MAXAVG) all eliminate the sawtooth pattern, with CU-MINMAX achieving the highest worst-case RankMe scores and CU-MAXAVG the highest average scores across the magnification range. Figure~\ref{fig:rep_quality_combined}B provides a numerical comparison of the RankMe scores to the accumulated training signal $S(y)$ (see~Equation~\ref{eq:domain_adaption}), showing strong alignment between our theoretical predictions and the empirical measurements in form of the normalized RankMe scores of the different strategies.

An analysis of the embedding space of the discrete uniform (DU) sampling model provides further insights into how the model represents the magnifications sampled at training time. Figure~\ref{fig:rep_quality_combined}C shows cosine similarities between embedding centroids computed at each standard magnification. Patches from similar magnifications cluster more closely, while larger magnification differences correspond to greater separation. This organization is also visible in t-SNE projections (Figure~\ref{fig:rep_quality_combined}D), where patches group strongly by magnification. Together, these observations indicate that magnification acts as a continuous organizing dimension in the learned representation space. Embedding space visualizations for other models provided in \ref{apd:embedding_space_visus}.

In summary, our analysis shows that established sampling strategies yield suboptimal cross-magnification representation quality. In contrast, the sampling methods derived from our framework yield a smooth representation and outperform naive continuous magnification sampling.


\subsection{Assessing Multi-Scale Downstream Performance}

\begin{figure*}[t!]
\centering
\includegraphics[width=\textwidth]{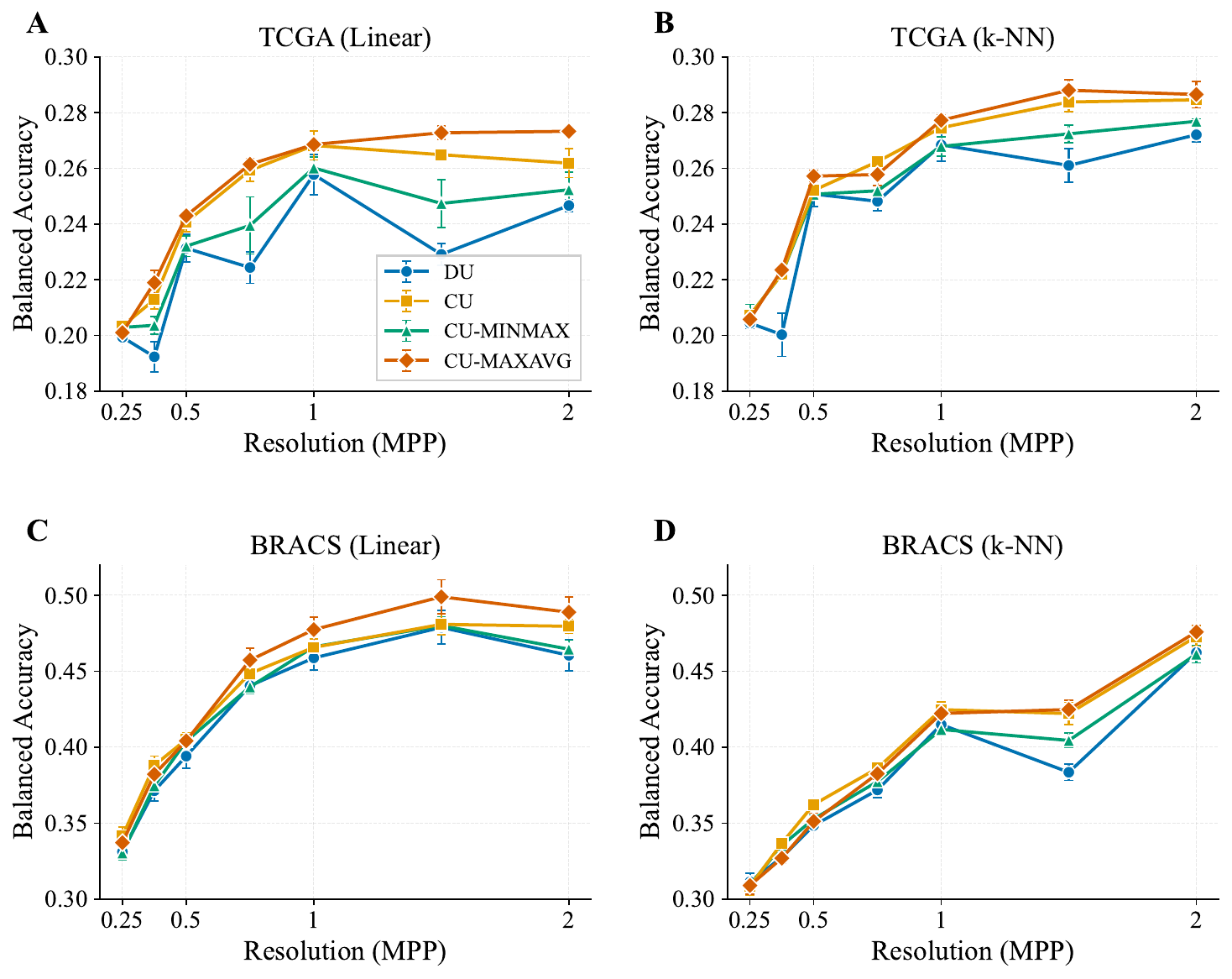}
\caption{Multi-scale classification performance across benchmarks. (A) TCGA-MS with linear classifier. (B) TCGA-MS with a k-NN classifier. (C) BRACS-MS with linear classifier. (D) BRACS-MS with k-NN classifier. All panels show balanced accuracy across resolutions (MPP). Error bars indicate the standard error of the mean over three training seeds. 
}
\label{fig:multiscale_comparison}
\end{figure*}

To evaluate whether the patterns observed in the representation space using the RankMe profiles translate to downstream task performance, we assess all multi-scale models on the TCGA-MS and BRACS-MS benchmarks using linear and k-NN classification in a 5-fold cross-validation stratified by patient and class (Figure~\ref{fig:multiscale_comparison}). We include the results in tabular form in \ref{apd:downstream_results_vits} and detail the evaluation protocol and hyperparameters in \ref{apd:benchmark_details}.

Continuous sampling strategies consistently outperform discrete uniform sampling (DU) at intermediate magnifications across benchmarks and evaluation methods. On TCGA-MS, discrete sampling shows clear performance dips at intermediate scales: CU improves over DU by 3.6 percentage points at 1.5 mpp and 3.5 pp at 0.75 mpp, with CU-MAXAVG achieving gains of 4.4 pp at 1.5 mpp. On BRACS-MS, the same pattern holds with k-NN classification, where CU improves over DU by 3.9 pp at 1.5 mpp. The effect is weaker with linear classification on BRACS-MS, where continuous and discrete strategies perform comparably at intermediate scales. While this variation suggests that how embedding space weaknesses manifest in downstream performance can depend on the task, it is clear that continuous sampling either matches or substantially improves upon discrete sampling at intermediate magnifications.

Among the continuous sampling strategies, CU-MAXAVG consistently outperforms CU, particularly at lower magnifications. On BRACS-MS with linear classification, the gains are 1.8 pp at 1.5 mpp and 0.9 pp at 2.0 mpp. On TCGA-MS, CU-MAXAVG exceeds CU by 0.8 pp at 1.5 mpp and 1.2 pp at 2.0 mpp. At high magnifications (0.25, 0.5 mpp), differences between CU and CU-MAXAVG are minimal and within standard error. These observations align precisely with the predicted representation quality from our domain adaptation framework (Figure~\ref{fig:rep_quality}).  Surprisingly, CU-MINMAX does not achieve better performance at boundary magnifications despite showing improved worst-case RankMe scores (Figure~\ref{fig:rep_quality_combined}). One explanation is that our benchmarks focus on cancer subtyping tasks that may not require the fine-grained cellular detail visible only at high magnifications. Mitotic counting or microorganism detection might better reveal the benefits of uniform performance across scales. This highlights that while the RankMe metric and our domain adaptation framework successfully predict several phenomena, the relationship between representation quality and downstream performance remains task-dependent, suggesting directions for future research.

\subsection{Evaluating Pathology Foundation Models}

\begin{figure*}[!htbp]
\vspace{-20mm} 
\centering
\begin{minipage}[t]{0.45\textwidth}
    \centering
    \vspace{0.5mm}
    \includegraphics[width=\textwidth]{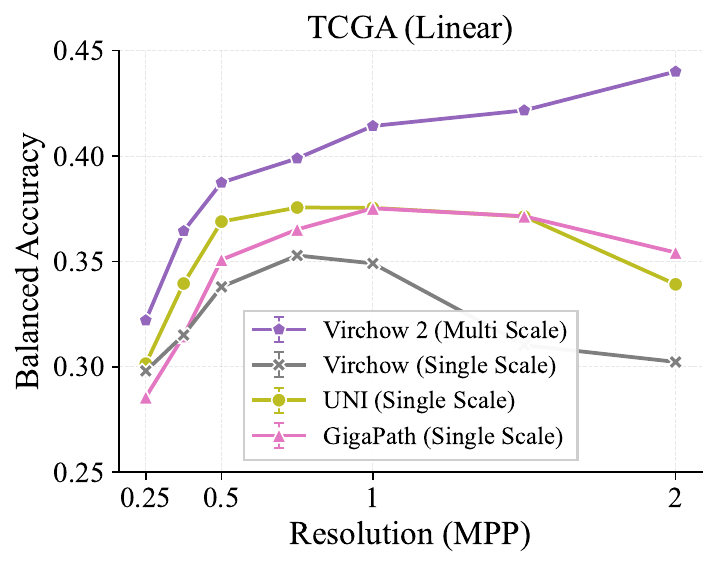}
\end{minipage}
\hfill
\begin{minipage}[t]{0.45\textwidth}
    \centering
    \vspace{0.5mm}
    \includegraphics[width=\textwidth]{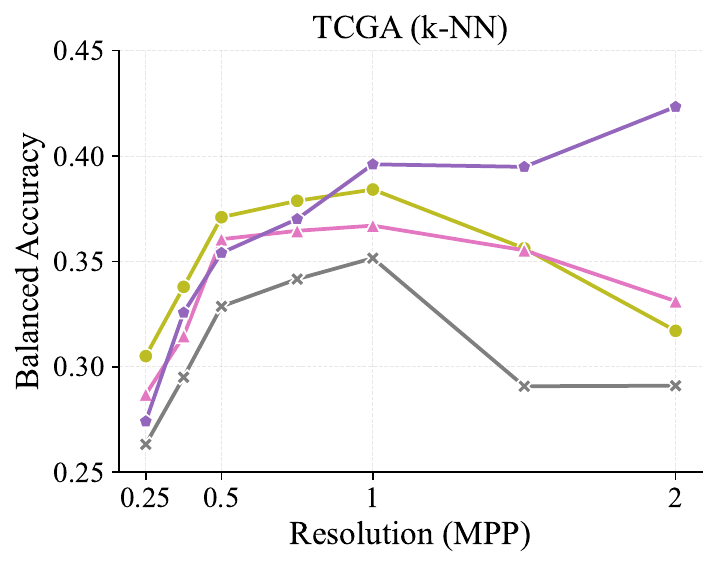}
\end{minipage}

\vspace{0.5em}

\begin{minipage}[t]{0.45\textwidth}
    \centering
    \vspace{0.5mm}
    \includegraphics[width=\textwidth]{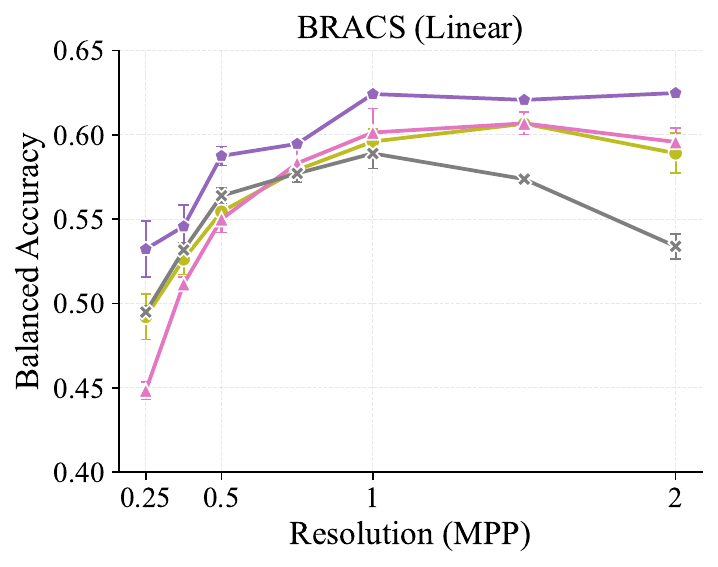}
\end{minipage}
\hfill
\begin{minipage}[t]{0.45\textwidth}
    \centering
    \vspace{0.5mm}
    \includegraphics[width=\textwidth]{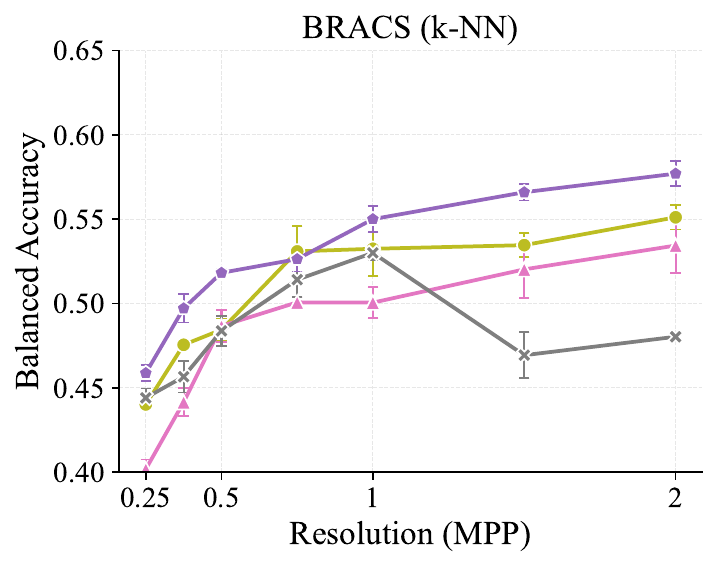}
\end{minipage}

\begin{minipage}[t]{0.45\textwidth}
    \centering
    \vspace{0.5mm}
    \includegraphics[width=\textwidth]{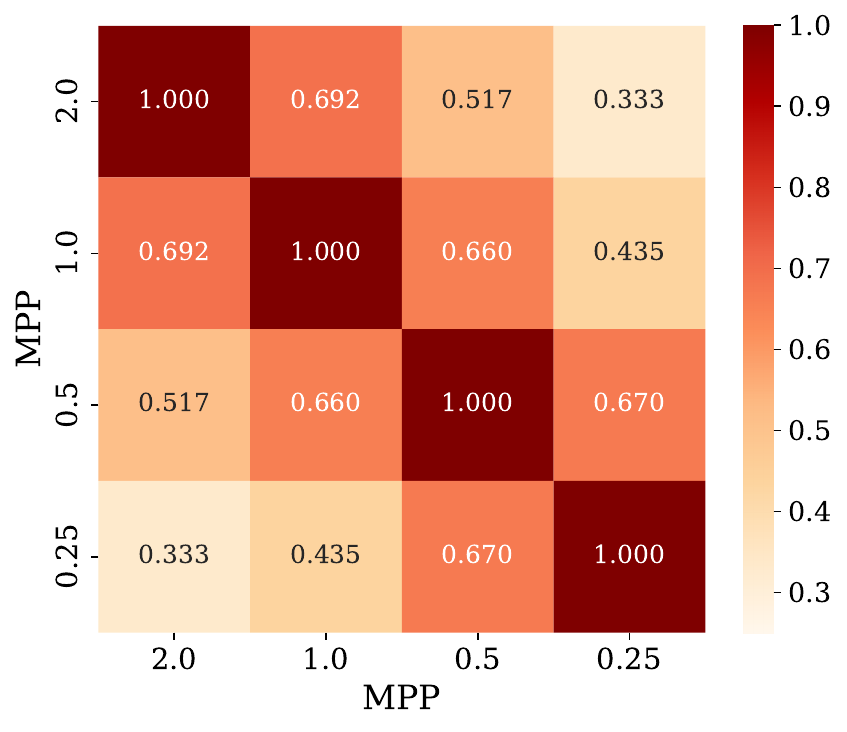}
\end{minipage}
\hfill
\begin{minipage}[t]{0.45\textwidth}
    \centering
    \vspace{0.5mm}
    \includegraphics[width=\textwidth]{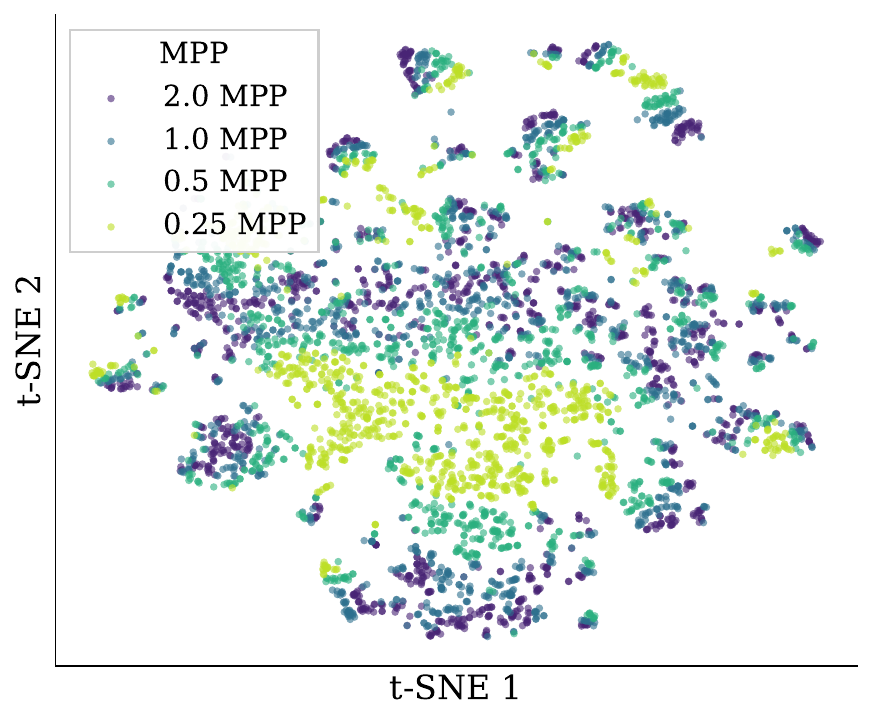}
\end{minipage}

\caption{Foundation model evaluation across magnification scales. (Top, Middle) Classification performance on TCGA (top) and BRACS (middle) via linear probe (left) and k-NN (right). Single-scale models degrade at extreme magnifications, particularly at 2.0 mpp, while Virchow2 (multi-scale) maintains robust performance across the spectrum. Nevertheless, performance dips at intermediate magnifications remain evident across both benchmarks. (Bottom) Virchow2 embedding space: cosine similarities between magnification centroids (left) and t-SNE projection (right). Virchow2 exhibits a magnification-dependent embedding structure consistent with our controlled experiments, though with less distinct stratification. }
\label{fig:FM_comparison_combined}
\end{figure*}

To investigate whether the performance profiles predicted by our domain adaptation framework extend to state-of-the-art foundation models, we evaluate a subset of publicly available models with disclosed training protocols. To get a clean signal and mitigate unwanted sources of variation, we exclude models trained on TCGA as we use it for our analysis and/or models that start training from pretrained imagenet checkpoints. Specifically, we compare Virchow 2~\citep{zimmermann_virchow2_2024}, a multi-scale model trained on 0.25, 0.5, 1.0 and 2.0 mpp, against three single-scale models trained predominantly at 0.5 mpp: Virchow~\citep{vorontsov_foundation_2024}, Prov-GigaPath~\citep{xu2024gigapath}, and UNI~\citep{chen_towards_2024}. These models range from ~300M to 1.1B parameters and are trained on orders of substantially more data than our controlled experiments. The results are depicted in Figure~\ref{fig:FM_comparison_combined}. We provide analysis of additional public foundation models (Phikon-v2 ~\citep{filiot2024phikon}, H-Optimus-0 ~\citep{hoptimus0}, Atlas ~\citep{alber_atlas_2025}, Uni-V2 ~\citep{chen_towards_2024}) in ~\ref{apd:fm-analysis} as a resource for the community.

\paragraph{Single-Scale Models} Consistent with our theoretical predictions and controlled experiments, single-scale models exhibit performance degradation at magnifications distant from their training scale (Figure~\ref{fig:FM_comparison_combined}). On both TCGA-MS and BRACS-MS, UNI, Prov-GigaPath, and Virchow show progressive performance drops as evaluation magnification moves away from 0.5 mpp, with the most pronounced degradation at 2.0 mpp. This smooth decline mirrors the patterns observed in our controlled single-scale ViT-S experiments (Figure~\ref{fig:vits_rankme_ss}). At 2.0 mpp on TCGA-MS, these models underperform the  multi-scale model Virchow2 by up to 12 percentage points, illustrating how single-scale or narrowly focused training can create substantial weaknesses at non-training magnifications.

\paragraph{Multi-Scale Model} Virchow 2, trained across four magnifications (0.25, 0.5, 1.0, and 2.0 mpp), maintains more robust performance across the spectrum. It substantially outperforms single-scale models at boundary magnifications, particularly at 2.0 mpp. However, consistent with our theoretical predictions for discrete uniform sampling, Virchow 2 exhibits performance dips at intermediate magnifications on both benchmarks. We linear interpolated between adjacent standard magnifications to estimate the effect size and find that largest gaps occur on TCGA-MS with KNN classification at 1.5 mpp (-1.49 pp) and on BRACS-MS with linear classification at 0.375 and 0.75 mpp (-1.42 and -1.13 pp respectively). While the weaknesses are less pronounced than in our controlled ViT-S experiments, likely due to larger model capacity and more diverse training data, it confirms that SOTA models are magnification sensitive and continuous sampling has the potential to improve them.

\section{Discussion}


Foundation models are increasingly deployed in digital pathology and enable practitioners to build powerful task-specific downstream applications with limited amounts of data. Nevertheless, critical limitations remain, and recent work has identified issues like susceptibility to technical artifacts \citep{kömen2025robustfoundationmodelsdigital, filiot2025distilling} and interpretability challenges \citep{Klauschen2024,kauffmann2025clever_hans}.
Our study contributes a novel perspective on foundation model performance and identifies magnification sampling as an additional important design decision that has been largely overlooked.

Since foundation models are used in the context of a wide range of disease pathologies whose characteristics are visible on different scales, we argue that computational tools must provide reliable performance across magnifications. For example, identifying isolated tumor cells in lymph nodes may require cellular resolution (0.25--0.5 mpp), while detecting a lymphoma may require an architectural overview (1.0--2.0 mpp). To investigate this, we framed magnification sampling as a domain adaptation problem (Section~\ref{sec:theoretical_framework}) and complemented it with empirical measurements of dimensional collapse and novel multi-scale benchmarks (Section~\ref{sec:ds_ms}). Using these tools, we demonstrated that magnification sampling meaningfully influences representation quality, exposed flaws of established discrete sampling approaches, and proposed continuous sampling strategies to address these shortcomings (Section~\ref{sec:experiments}). We anticipate that this will have broad implications for the community.

 

First, our results indicate that foundation model performance rankings can shift substantially depending on the evaluated magnification. A model that excels on a breast cancer subtyping benchmark at 0.5 mpp may underperform at the coarser magnifications relevant for assessing tissue architecture. Model selection without magnification-specific evaluation is therefore unreliable, and we propose that practitioners selecting foundation models for deployment should evaluate performance at their target magnifications rather than relying solely on task-specific benchmark rankings.

Second, many current foundation models perform well on 0.5 mpp, which is also the chosen magnification of many public benchmarks for evaluation. However, we showed that this over-optimization can lead to suboptimal performance for clinical practitioners who do not rely exclusively on this magnification. We hope that creating awareness of this issue leads to future development of practices that reduce overfitting on existing benchmarks, and that the community contributes more multi-scale benchmarks for foundation model evaluation. Furthermore, we highlight that there is an opportunity to build better foundation models for coarse magnifications (e.g., from 1.0-2.0 mpp), where many existing models underperform despite the importance of coarse tissue features for many clinical tasks.

Third, in histopathology, tissue is commonly analyzed at discrete magnifications. Yet, digital pathology enables integrating information from continuous magnification levels, and our work demonstrates the potential benefits of this perspective. Specifically, our continuous magnification sampling strategies clearly outperformed discrete sampling, even on common discrete magnification levels. We anticipate that this principle extends beyond pretraining. Slide-level representations that aggregate information across scales, task-specific fine-tuning, and interactive diagnostic tools could all benefit from treating magnification as a continuous rather than a discrete variable.

We proposed three tools for analyzing foundation model performance across magnifications, each coming with advantages and limitations. (i)~Our theoretical framework enables modeling the effects of magnification sampling and deriving principled alternatives to uniform sampling without requiring real-world data. Yet, it may not fully capture the complexity of large-scale pretraining. (ii)~Profiling embedding spaces with the RankMe metric only requires unlabeled data and is independent of downstream task characteristics. However, it may capture both relevant and irrelevant features, and may not perfectly predict downstream model performance. (iii)~Benchmarking multi-scale downstream model performance exposes magnification effects in practically relevant tasks. However, it requires sufficiently large labeled datasets and is influenced by task-specific magnification effects (e.g., a to-be-predicted disease may not be visible at 0.25 mpp). Interestingly, some trends were observed in all analytical layers, such as performance degradation at intermediate magnifications for discrete sampling models. Together, the tools provided in this work enable a comprehensive assessment and improvement of foundation model performance across magnifications.

Some limitations of our work motivate potential future research. First, our similarity kernels provide useful approximations, but do not capture the full complexity of how visual features relate across magnifications. Second, our downstream evaluation focuses on cancer subtyping tasks; the clinical impact of magnification performance may vary for other diagnostic applications, such as mitotic counting or microorganism detection. Third, although we validated our principled magnification strategies through large-scale pretraining runs, we could not yet apply them at data and model scales comparable to those of state-of-the-art foundation models, due to the immense computational budget that this requires. Fourth, while we treat magnification sampling in isolation, practical data curation must balance multiple sources of variation and diversity, including tissue types, staining protocols, and scanner characteristics. Extending our framework to jointly reason about these dimensions represents a promising direction for future work.

\newpage

\bibliographystyle{plainnat}
\bibliography{references.bib}


\appendix

\section{Background and Related Work}

\paragraph{Vision foundation models in digital pathology}
Microscopic visual data in digital pathology consists of gigapixel whole-slide images (WSIs) that capture human tissue at high resolution. Because of their size, WSIs are typically divided into a grid of uniformly sized image patches. Patch-level foundation models are then trained using large-scale self-supervised learning frameworks such as DINOv2 \citep[e.g.,][]{chen_towards_2024, dippel_rudolfv_2024, zimmermann_virchow2_2024, vorontsov_foundation_2024, alber_atlas_2025}. The resulting patch embeddings capture fine-grained local tissue morphology and can be used for a variety of downstream applications, including cell segmentation and classification \citep[][]{horst2024cellvit, luscher_2025}, rare disease detection \citep[][]{dippel_AD_nejm}, and other slide-level or region-level tasks \citep[][]{Hense2025}.
More recently, advanced aggregation strategies built on top of these patch-level encoders have enabled the development of whole-slide foundation models that aim to represent WSIs directly \citep[e.g.,][]{chief_wang2024, xu2024gigapath, titan_ding2025, feathermil_shao2025}. Foundation model performance has been probed and improved to generalize across tissues, scanners and hospital \citep[][]{carloni2025pathology,drexlin2025medi,subramanian2025starc}, but no work has investigated magnification robustness in detail. A preliminary version of the present work has appeared as a workshop paper \citep{moellers2025continuous}.

\paragraph{Multi-scale self-supervised training in digital pathology}
Training pathology foundation models has recently shifted from mono-scale learning \citep[][]{ctranspath_wang2022} to multi-scale training \citep[e.g.,][]{ciga_self_2022,zimmermann_virchow2_2024, alber_atlas_2025, grashei2025pathryoshkacompressingpathologyfoundation}. These multi-scale models typically train on patches sampled from four standard resolutions (0.25, 0.5, 1.0, and 2.0 $\mu$m per pixel), with recent implementations adopting uniform sampling across these magnifications \citep[e.g.,][]{ai_towards_2024, karasikov2025trainingstateoftheartpathologyfoundation}. Beyond sampling strategies, researchers have explored architectural mechanisms for multi-scale learning, including positional encodings that capture relative magnification differences between patches \citep[]{zimmermann_adapting_2024} and masked autoencoder objectives that reconstruct patch regions of varying sizes \citep{juyal_pluto_2024}. Complementary work in vision-language models has explored how multi-scale training on standard magnifications improves responses to scale-dependent visual prompts \citep{albastaki2025multi}.
Despite these advances, the choice and effects of sampling multi-scale data have not yet been thoroughly investigated.

\paragraph{Multi-scale Benchmarks and Applications in Digital Pathology}\label{rel_work:ms_benchmarks}
Multi-scale analysis is essential for many clinical applications, such as multi-magnification image retrieval \citep{Rasoolijaberi_2022} and cross-scale cancer classification \citep{deng2024crossscalemultiinstancelearningpathological}. To assess performance, several multi-magnification benchmarks have been introduced \citep[e.g.,][]{breakhis, tcga-ut, vaidya2025molecular}, but they do not cover all common scanner resolutions (0.25, 0.5, 1.0, and 2.0 mpp) and cannot assess model behavior at intermediate magnifications. To address this gap, we introduce TCGA-MS and BRACS-MS, which enable a comprehensive evaluation across the magnification spectrum.

\section{Training Details}\label{apd:training_details}

For training, we adapt the DinoV2 framework \citep{oquab2024dinov2learningrobustvisual} and train a student network $f_{\theta_s}(\mathbf{X})$ and a teacher network $f_{\theta_t}(\mathbf{X})$  for 60.000 iterations with a batch size of 320, with $\mathbf{X}\in R^{224x224}$. We extract two global crops $\mathbf{X}_{g}\in R^{224x224}$ and eight local crops $\mathbf{X}_{l}\in R^{98x98}$  from a larger source patch $\mathbf{X}_{s}\in R^{256x256}$. Furthermore, we create masked versions of the global crops $\mathbf{X}_{g_{m}}$. The objective is then a combination of the Dino loss $\mathcal{L}_{Dino}$, the Ibot loss $\mathcal{L}_{ibot}$ and the Koleo loss $\mathcal{L}_{Koleo}$ \citep{caron2021emergingpropertiesselfsupervisedvision, zhou2021ibot, sablayrolles2018spreading} that together encourage image-level distillation between the local and global crops and patch-level reconstruction between the masked and unmasked ones \cite{oquab2024dinov2learningrobustvisual}. As architecture, we use small vision transformers (ViT-S) and use the teacher network to create the final embeddings. For $N$ patches this results in an embedding matrix $\mathbf{Z} \in R^{Nx384}$. We use the standard DinoV2 hyperparameters and train with a base learning rate of 0.001 and a weight decay cosine schedule from 0.04 to 0.2. The scale ranges for the resizing of global and local crops are set to [1, 0.35] and [0.35, 0.05] respectively.

\section{Domain Adaptation Framework}  \label{apd:domain_adaption_framework}

In this section, we provide further details on how we find the optimal distributions based on the domain adaptation framework, discuss the impact of kernel choice, provide a proof for proposition~\ref{prop:mag_prototypes} and include additional discussions on the Kernel choice. 

\paragraph{Max-Average Optimization with Entropy Regularization}
To find the optimal distribution $p^*(x)$ for the max-average optimization problem in Equation~\ref{def:max_avg} for the information based kernel, we note that we can use the method of lagrange multipliers. This yields an analytical solution of the form:

\begin{equation}
p^*(x) = \frac{\exp(\bar{K}(x)/\lambda)}{\int_{0.25}^{2} \exp(\bar{K}(x)/\lambda) dx}
\end{equation}

where $\bar{K}(x) = \int_{0.25}^{2}  K(x,y) \, dy$ is the transfer potential discussed in Section \ref{sec:transfer_potential}.

\paragraph{Max-Min Optimization.}
To solve the max-min problem that maximizes worst-case representation quality as defined in Equation~\ref{def:max_min}, we discretize the problem on a grid of 1000 points spread uniformly between 0.25 and 2.0 mpp. We then solve
\begin{equation}
\max_{\mathbf{p}, t} \; t \quad \text{s.t.} \quad \mathbf{K}\mathbf{p} \geq t \cdot \mathbf{1}, \; \mathbf{p} \geq \mathbf{0}, \; \mathbf{1}^\top \mathbf{p} = 1
\end{equation}
 with $K_{ij}=K(x_{i},x_{j})$
 using the CVXPY package \citep{diamond2016cvxpy,agrawal2018rewritingcvxpy}.

\paragraph{Proof of Proposition \ref{prop:mag_prototypes} (Magnification Prototypes)}\label{proof:mag_prototypes} The proof follows the rules of calculus and simply takes the derivative of the transfer kernel function and sets it to $0$ to find the extreme values. For a symmetric kernel $K(x,y) = f(|x-y|)$ with $f$ monotonically decreasing, the transfer potential is:
\[
\bar{K}(x) = \int_{a}^{b} f(|x-y|) \, dy.
\]

To be able to differentiate, we split at $x$ to get rid of the absolute value and substitute $u = x - y$ and $v = y - x$ respectively:
\[
\bar{K}(x) = \int_{0}^{x-a} f(u) \, du + \int_{0}^{b-x} f(v) \, dv = F(x-a) + F(b-x),
\]
We can then take the derivative of this as:
\[
\frac{d\bar{K}}{dx} = f(x-a) - f(b-x).
\]

Setting it to zero we find that our extrema lies at: 
\[
 f(x-a) = f(b-x).
\]

as $f$ is strictly decreasing we need $x-a = b-x$ and thus $x = \frac{a+b}{2}$. The second derivative $\frac{d^2\bar{K}}{dx^2} = f'(x-a) + f'(b-x) < 0$ confirms this is a maximum. 

\paragraph{Kernel Choice}

\begin{figure}[t!]
\centering
\includegraphics[width=0.6\textwidth]{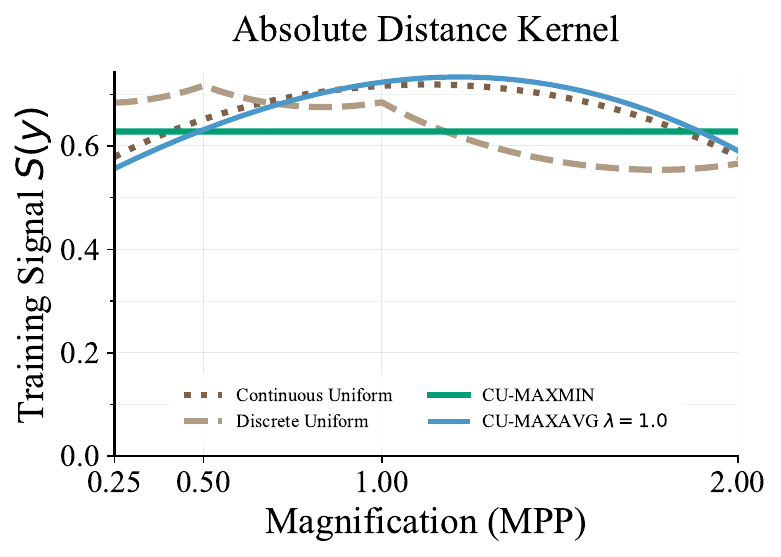}
\caption{Accumulated training signal $S(y)$ across magnifications under the absolute distance kernel. While the kernel correctly predicts degradation at intermediate magnifications for discrete sampling (DU), it underestimates the magnitude of this effect compared to empirical observations (cf.\ Figure~\ref{fig:rep_quality_combined}A), motivating our use of the information-based kernel for the primary analysis.}
\label{fig:rep_qual_abs_distance}
\end{figure}

Our theoretical framework admits multiple choices for the similarity kernel $K(x,y)$, and a natural question is whether our conclusions are sensitive to this choice. Figure~\ref{fig:rep_qual_abs_distance} shows the accumulated training signal $S(y)$ under the absolute distance kernel $K_{\text{abs}}(x,y) = \frac{1}{1 + |x-y|}$. While this kernel captures the general phenomenon of degradation at intermediate magnifications for discrete sampling, it predicts a less pronounced contrast between discrete and continuous strategies than we observe empirically. In particular, the dimensional collapse at intermediate magnifications evident in our RankMe analysis (Figure~\ref{fig:rep_quality_combined}A) is more accurately captured by the information-based kernel, which models the quadratic relationship between field-of-view overlap and magnification ratio. We therefore adopt the information-based kernel for our primary analysis.

Nevertheless, the qualitative properties of the optimized sampling distributions remain consistent across kernel choices. For both kernels, the max-min objective yields distributions that oversample boundary magnifications to compensate for their reduced transfer potential (Figure~\ref{fig:max_min_dist}), while the entropy-regularized max-average objective produces distributions concentrated toward central magnifications. Moreover, the regularization parameter $\lambda$ in Equation~\ref{def:max_avg} provides an additional degree of freedom that can yield similar distributions across different kernels: increasing $\lambda$ for a kernel with sharper decay produces distributions comparable to those obtained with lower $\lambda$
for a kernel with broader support. This suggests that the fundamental trade-off between worst-case and average-case optimization that we identify is robust to the specific functional form of the kernel

\section{Evaluation Protocol \& Benchmark Details}  \label{apd:benchmark_details}

\subsection{Evaluation Protocol}\label{apd:eval_protocol}

\paragraph{Data Splits} Both benchmarks use 5-fold cross-validation stratified by both patient and class to prevent data leakage and ensure balanced class representation across folds. For each fold, data is partitioned into 60\% training, 20\% validation, and 20\% test sets. To ensure robust performance estimates for BRACS-MS, we repeat cross-validation over three random seeds as the dataset is comparably small (10,773 patches) and evaluation time allows for it.

\paragraph{Embedding Extraction} For all models, we extract embeddings using a concatenation of the [CLS] token and mean-pooled patch tokens (denoted \texttt{cls+mean}). Images are normalised using model-specific parameters where available, or ImageNet statistics otherwise.

\paragraph{Classifiers} We evaluate using two classifiers on frozen embeddings:
\begin{itemize}
    \item \textbf{Logistic Regression:} L2-regularised multinomial logistic regression with balanced class weights. The regularisation parameter $C$ is tuned with a grid search over $C \in \{10^{-4}, 10^{-3}, \ldots, 10^{4}\}$, selecting the value that maximises balanced accuracy on the validation set.
    \item \textbf{k-Nearest Neighbours (k-NN):} Cosine distance-based k-NN classifier. The number of neighbours $k$ is tuned over $k \in \{1, 3, 5, 10, 20, 40, 80\}$, we select the value that maximises balanced accuracy on the validation set.
\end{itemize}

\paragraph{Training Data \& Evaluation} Classifiers are trained on patches from the four standard magnifications (0.25, 0.5, 1.0, 2.0 mpp). For TCGA-MS, we sample 75 patches per slide at each training magnification. Test performance is evaluated separately at each of the seven magnifications. This design mirrors common real-world set up where labels are collected on standard magnifications and directly tests wether models can maintain performance in the blind sports where we observe dimensional collapse. We report balanced accuracy to account for class imbalance, with mean computed across folds (and seeds for BRACS-MS).

\subsection{TCGA-MS: Morphological Subtypes}

Table~\ref{tab:tcga_subtypes} lists the 16 morphological subtypes included in TCGA-MS. For each subtype, we selected eight whole-slide images from TCGA, yielding 128 WSIs in total.

\begin{table}[t]
\centering
\caption{Morphological subtypes included in TCGA-MS.}
\label{tab:tcga_subtypes}
\begin{tabular}{ll}
\toprule
\textbf{Morphological Subtype} & \textbf{Tissue Origin} \\
\midrule
Clear cell adenocarcinoma & Kidney \\
Chromophobe renal cell carcinoma & Kidney \\
Diffuse-type carcinoma & Stomach \\
Fibromyxosarcoma & Connective tissue \\
Fibrous histiocytoma & Connective tissue \\
Invasive carcinoma of no special type & Breast \\
Leiomyosarcoma & Connective tissue \\
Lobular carcinoma & Breast \\
Medullary carcinoma & Breast, Stomach \\
Metaplastic carcinoma & Breast \\
Mucinous adenocarcinoma & Breast, Stomach \\
Papillary adenocarcinoma & Stomach \\
Signet ring cell carcinoma & Stomach \\
Synovial sarcoma & Connective tissue \\
Tubular adenocarcinoma & Stomach \\
Undifferentiated sarcoma & Connective tissue \\
\bottomrule
\end{tabular}
\end{table}

\clearpage

\section{Analysis of Public Pathology FMs}  \label{apd:fm-analysis}

Using the benchmarks introduced in the paper, we provide additional results for publicly available pathology foundation models. In the main paper, we focused on models with disclosed training protocols that were not trained on TCGA to ensure a clean evaluation signal. Here, we extend our analysis to additional models, although we caution that interpretation is more difficult. Several of these models were trained on TCGA (which overlaps with our evaluation data), and many have of them have undisclosed or only partially disclosed training strategies. Combined with the fact that each model represents a single training run and differs in architecture, data composition, augmentation strategies, and other hyperparameters it is difficult to draw causal conclusions and the analysis should be considered descriptive.

\begin{figure}[!ht]
\centering
\begin{subfigure}[c]{0.35\textwidth}
    \includegraphics[width=\textwidth]{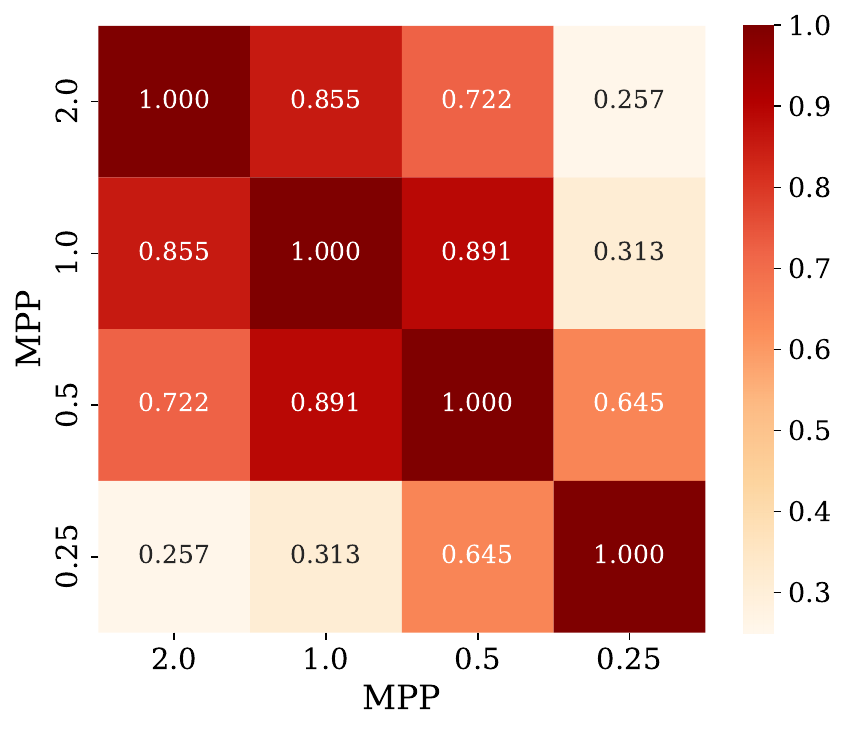}
\end{subfigure}
\hfill
\begin{subfigure}[c]{0.35\textwidth}
    \includegraphics[width=\textwidth]{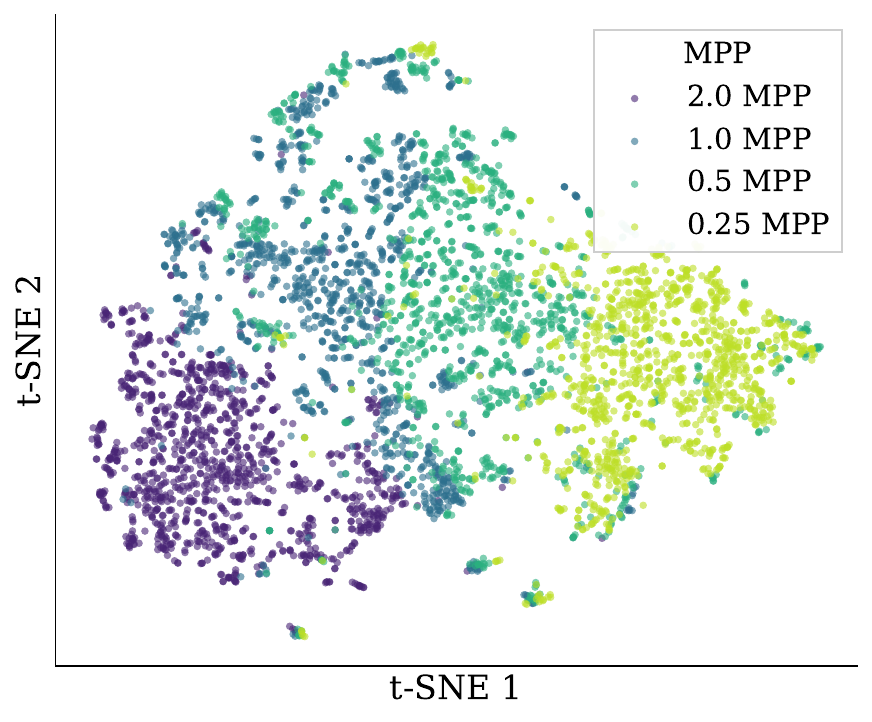}
\end{subfigure}
\caption{Virchow}
\label{apd:Virchow_heatmap}
\end{figure}

\begin{figure}[!ht]
\centering
\begin{subfigure}[c]{0.35\textwidth}
    \includegraphics[width=\textwidth]{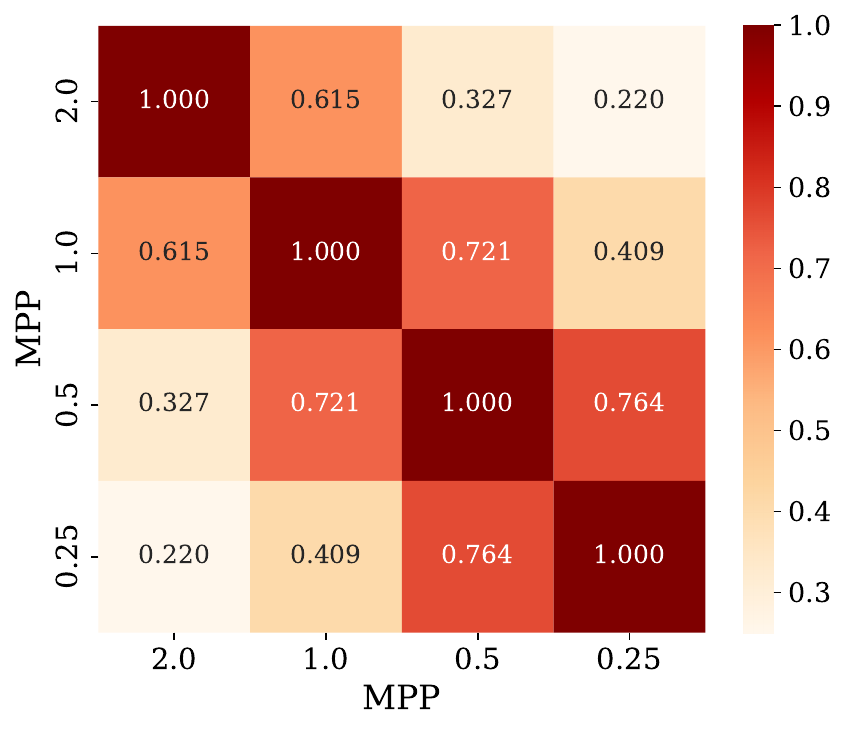}
\end{subfigure}
\hfill
\begin{subfigure}[c]{0.35\textwidth}
    \includegraphics[width=\textwidth]{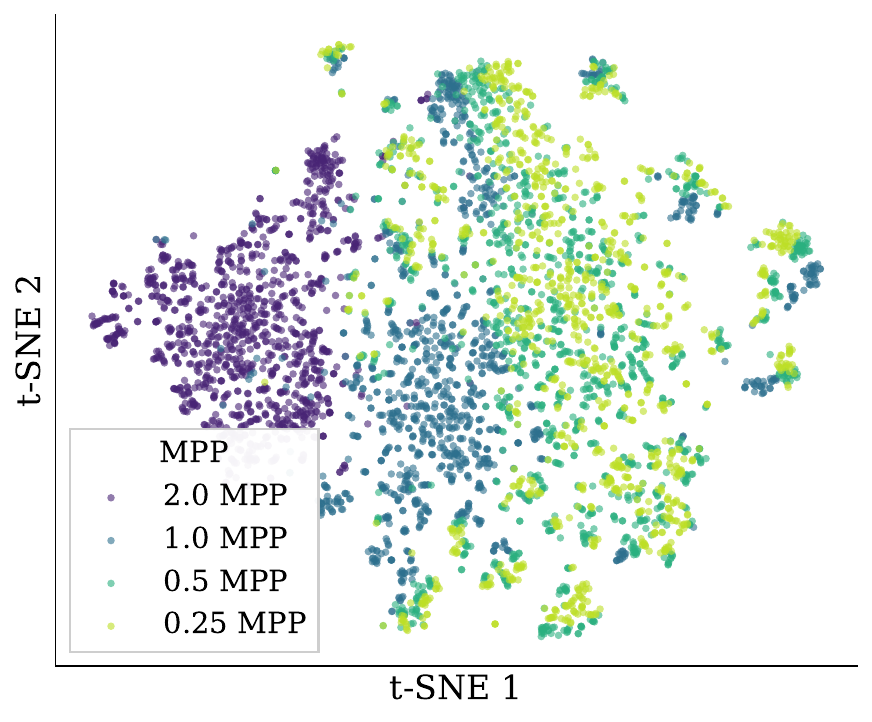}
\end{subfigure}
\caption{H-optimus-0}
\label{apd:Hoptimus_heatmap}
\end{figure}

Despite these caveats, we observe that magnification remains a significant organizing principle in the embedding spaces of all models, and performance variations across scales persist even in state-of-the-art systems (Figures~\ref{apd:Virchow_heatmap}--\ref{apd:Hoptimus_heatmap}). We present these results to characterize the current landscape and as a resource for practitioners selecting models for specific magnification requirements.

\subsection{Downstream Task Performance}
\begin{figure*}[!htbp]
\centering
\begin{minipage}[t]{0.45\textwidth}
    \centering
    \vspace{0.5mm}
    \includegraphics[width=\textwidth]{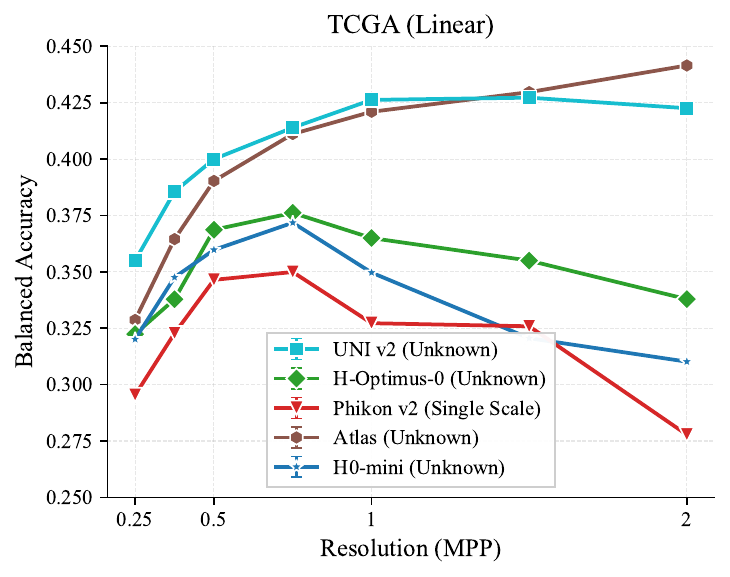}
\end{minipage}
\hfill
\begin{minipage}[t]{0.45\textwidth}
    \centering
    \vspace{0.5mm}
    \includegraphics[width=\textwidth]{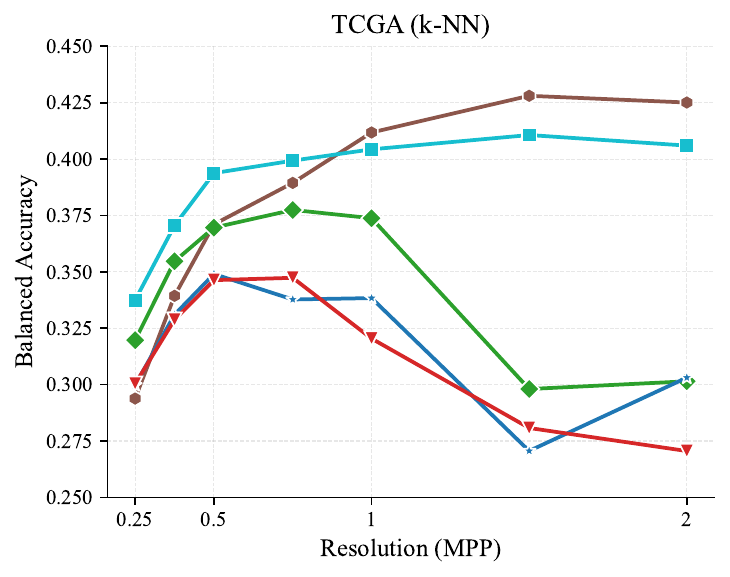}
\end{minipage}

\vspace{0.5em}

\begin{minipage}[t]{0.45\textwidth}
    \centering
    \vspace{0.5mm}
    \includegraphics[width=\textwidth]{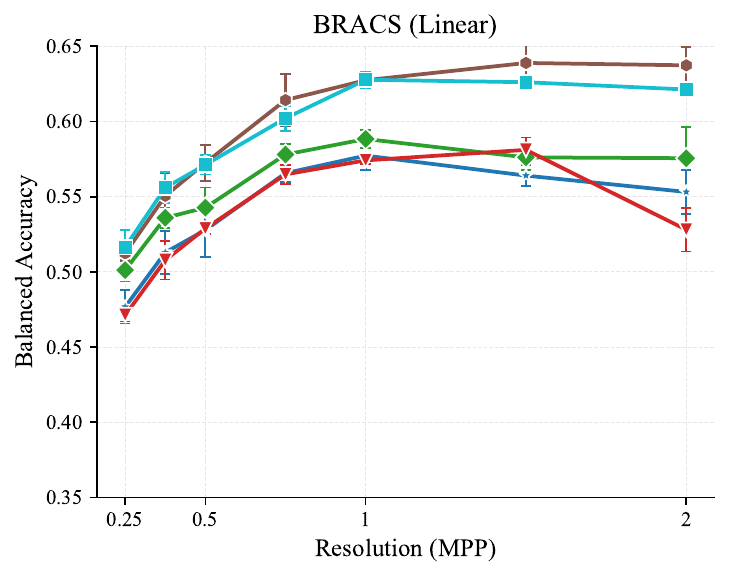}
\end{minipage}
\hfill
\begin{minipage}[t]{0.45\textwidth}
    \centering
    \vspace{0.5mm}
    \includegraphics[width=\textwidth]{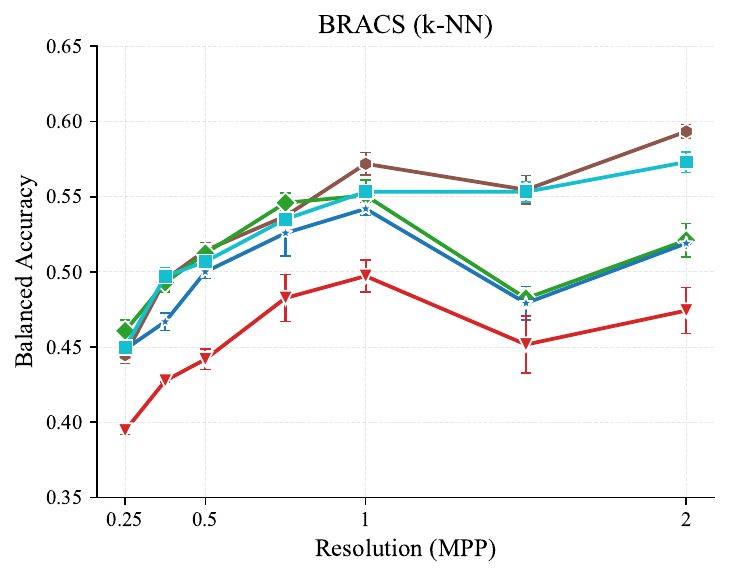}
\end{minipage}

\caption{Foundation model evaluation across magnification scales. }
\label{apd:Apd_FM_comparison_combined}
\end{figure*}

Figure~\ref{apd:Apd_FM_comparison_combined} presents classification performance across magnifications on TCGA-MS and BRACS-MS for both linear and KNN evaluation protocols. Despite the confounds discussed above, we observe that magnification-dependent performance variation persists across all models, confirming that magnification robustness remains a relevant consideration even for state-of-the-art systems.

\paragraph{Performance Profiles} Models exhibit diverse magnification profiles that likely reflect their training strategies. H-Optimus-0, Phikon-v2, and H0-mini show strong performance at higher magnifications (0.25--0.5 mpp) but exhibit substantial degradation as we move towards 2.0 mpp, consistent with single-scale or high-magnification-focused training. Among the strongest overall performers, UNI-v2 and Atlas show complementary profiles. UNI-v2 excels at high magnifications (0.25--0.5 mpp), particularly on TCGA-MS, while Atlas demonstrates stronger performance at lower magnifications (1.0--2.0 mpp). Notably, neither model exhibits the sawtooth pattern characteristic of discrete uniform sampling, though without disclosed training details the underlying cause remains unclear. This suggests that even top-performing models may have magnification-specific strengths, and model selection should consider the target application's magnification requirements.

\clearpage

\begin{figure}[!ht]
\centering
\begin{subfigure}[c]{0.35\textwidth}
    \includegraphics[width=\textwidth]{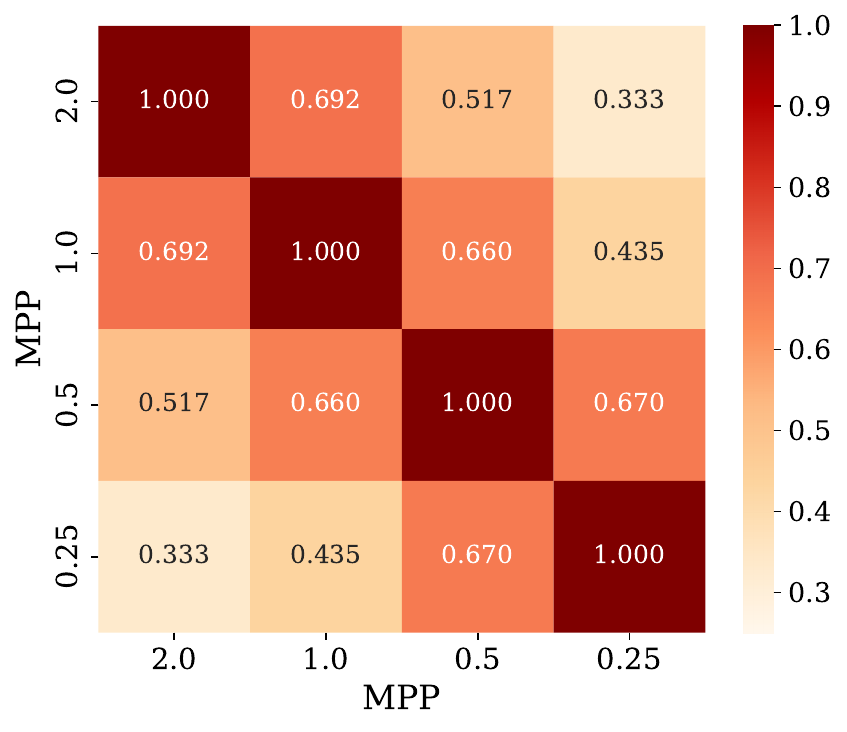}
\end{subfigure}
\hfill
\begin{subfigure}[c]{0.35\textwidth}
    \includegraphics[width=\textwidth]{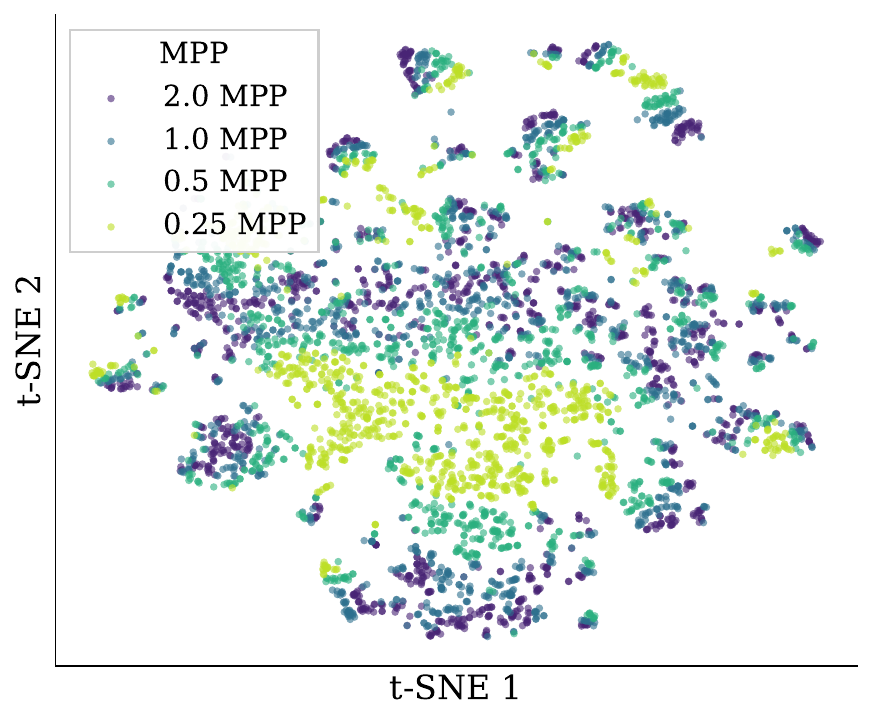}
\end{subfigure}
\caption{Virchow2}
\end{figure}

\begin{figure}[!ht]
\centering
\begin{subfigure}[c]{0.35\textwidth}
    \includegraphics[width=\textwidth]{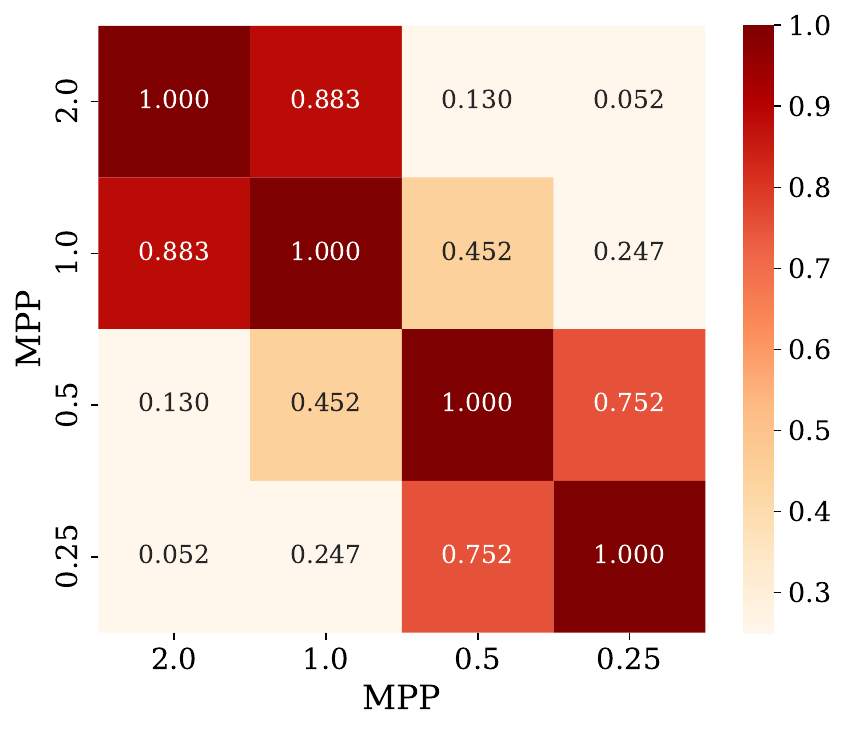}
\end{subfigure}
\hfill
\begin{subfigure}[c]{0.35\textwidth}
    \includegraphics[width=\textwidth]{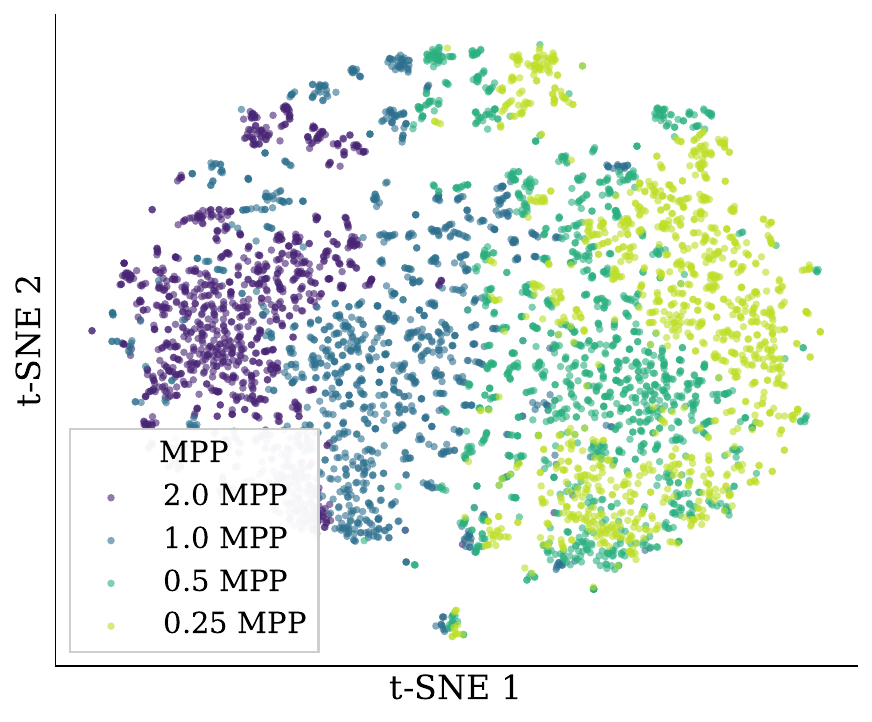}
\end{subfigure}
\caption{Prov-GigaPath}
\end{figure}

\clearpage

\begin{figure}[!ht]
\centering
\begin{subfigure}[c]{0.35\textwidth}
    \includegraphics[width=\textwidth]{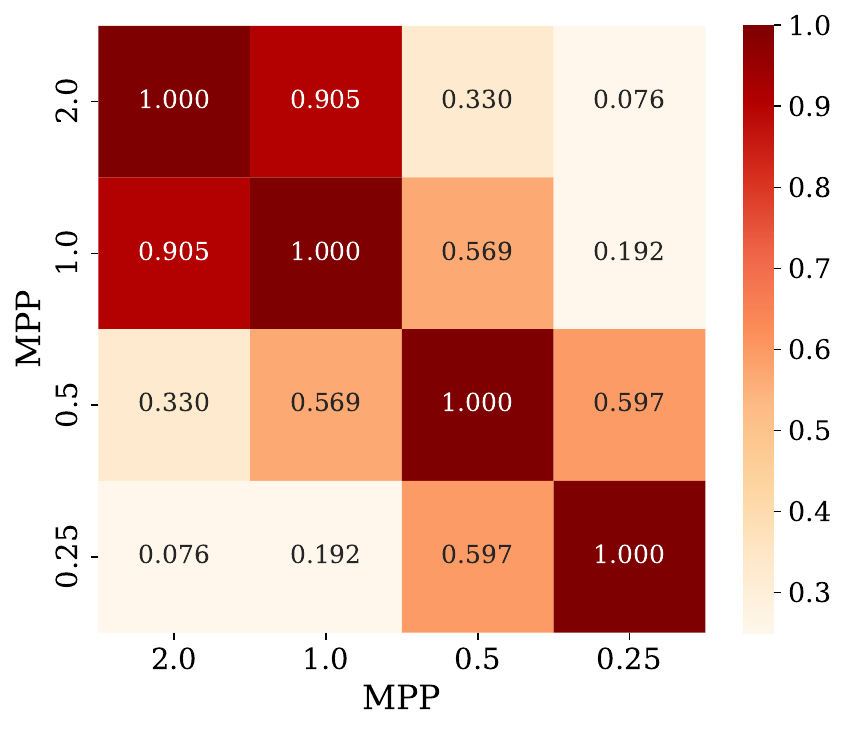}
\end{subfigure}
\hfill
\begin{subfigure}[c]{0.35\textwidth}
    \includegraphics[width=\textwidth]{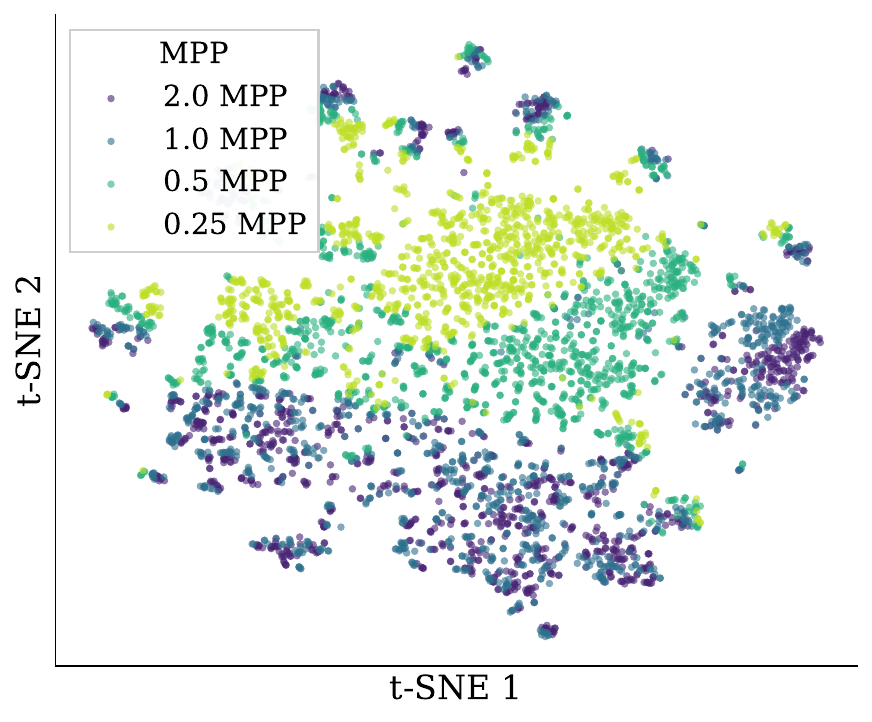}
\end{subfigure}
\caption{Uni-V2}
\end{figure}

\begin{figure}[!ht]
\centering
\begin{subfigure}[c]{0.35\textwidth}
    \includegraphics[width=\textwidth]{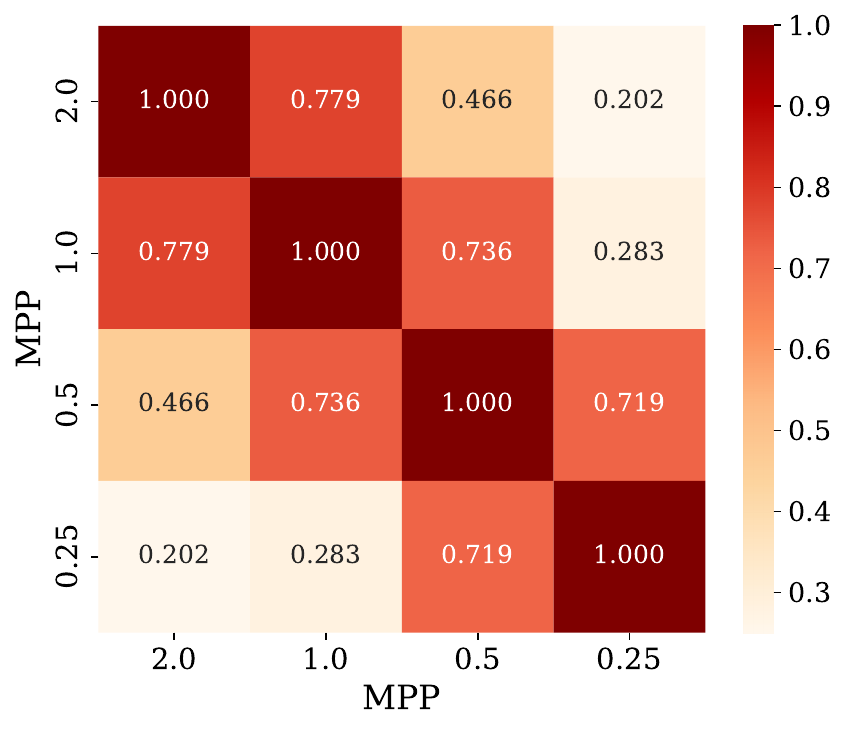}
\end{subfigure}
\hfill
\begin{subfigure}[c]{0.35\textwidth}
    \includegraphics[width=\textwidth]{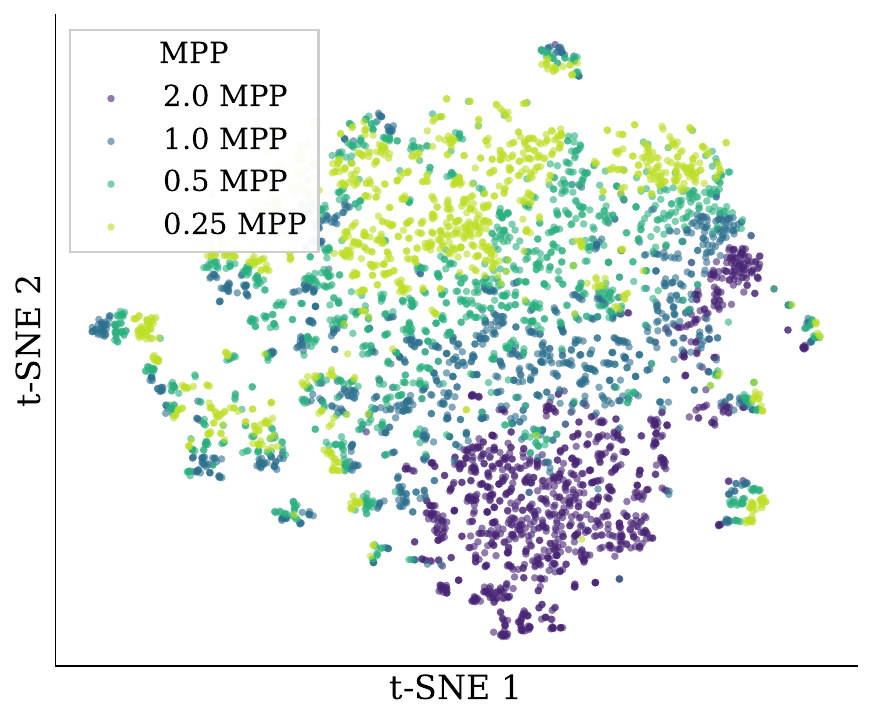}
\end{subfigure}
\caption{Uni}
\end{figure}

\begin{figure}[!ht]
\centering
\begin{subfigure}[c]{0.35\textwidth}
    \includegraphics[width=\textwidth]{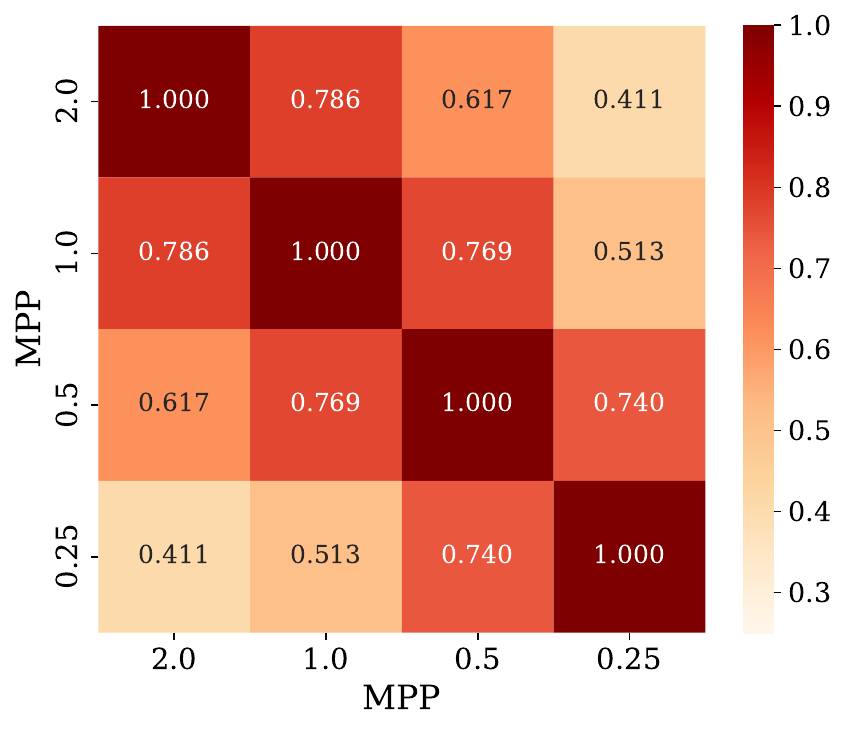}
\end{subfigure}
\hfill
\begin{subfigure}[c]{0.35\textwidth}
    \includegraphics[width=\textwidth]{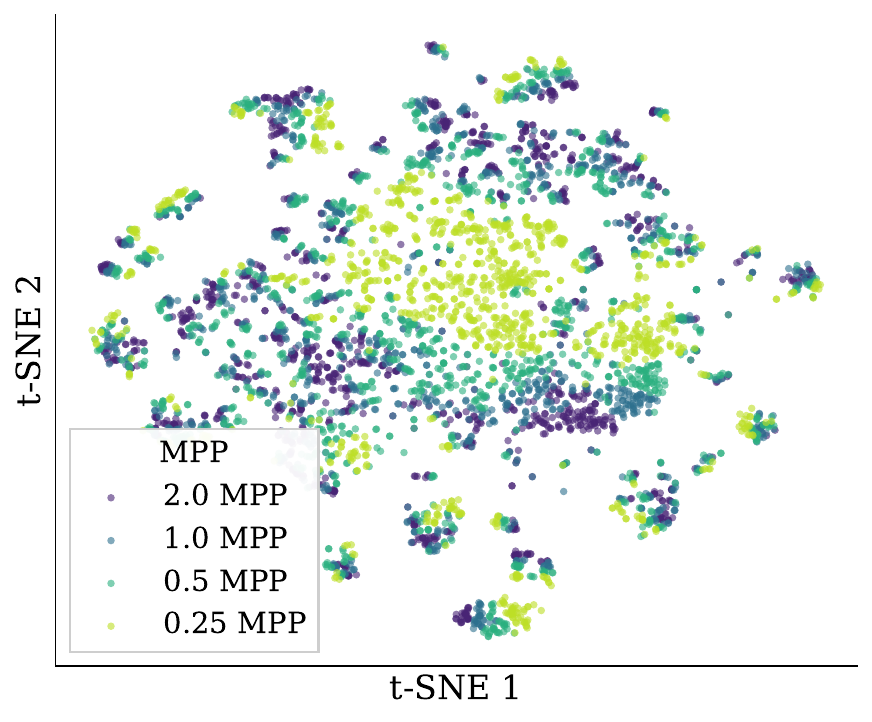}
\end{subfigure}
\caption{Atlas}
\end{figure}

\clearpage

\clearpage
\section{Embedding Space Visualization by Model} \label{apd:embedding_space_visus}

\begin{figure}[!ht]
\centering
\begin{subfigure}[c]{0.35\textwidth}
    \includegraphics[width=\textwidth]{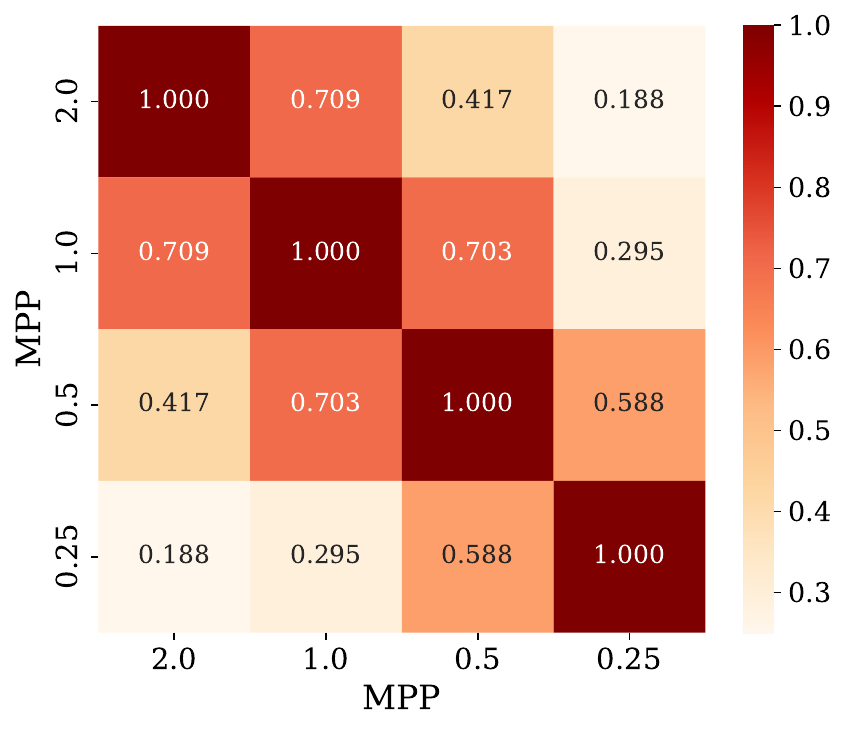}
\end{subfigure}
\hfill
\begin{subfigure}[c]{0.35\textwidth}
    \includegraphics[width=\textwidth]{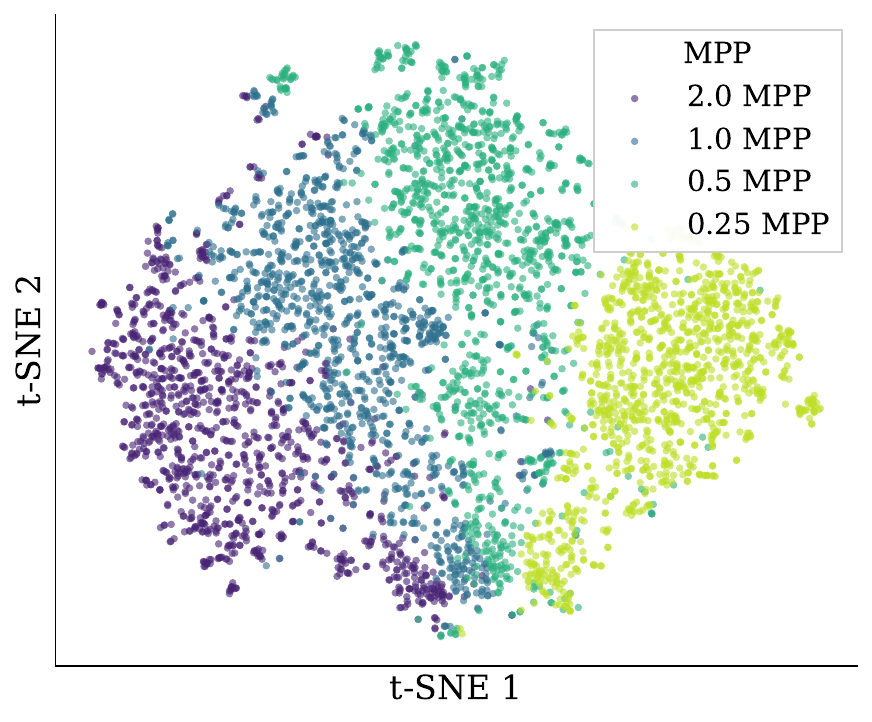}
\end{subfigure}
\caption{MMPP (DinoV2 AUG)}
\end{figure}

\begin{figure}[!ht]
\centering
\begin{subfigure}[c]{0.35\textwidth}
    \includegraphics[width=\textwidth]{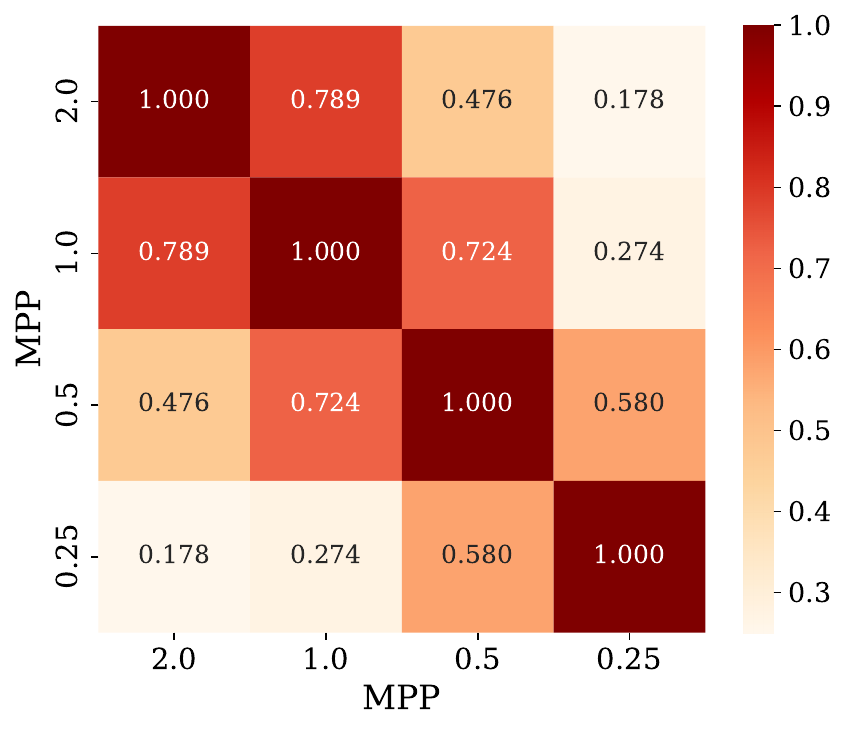}
\end{subfigure}
\hfill
\begin{subfigure}[c]{0.35\textwidth}
    \includegraphics[width=\textwidth]{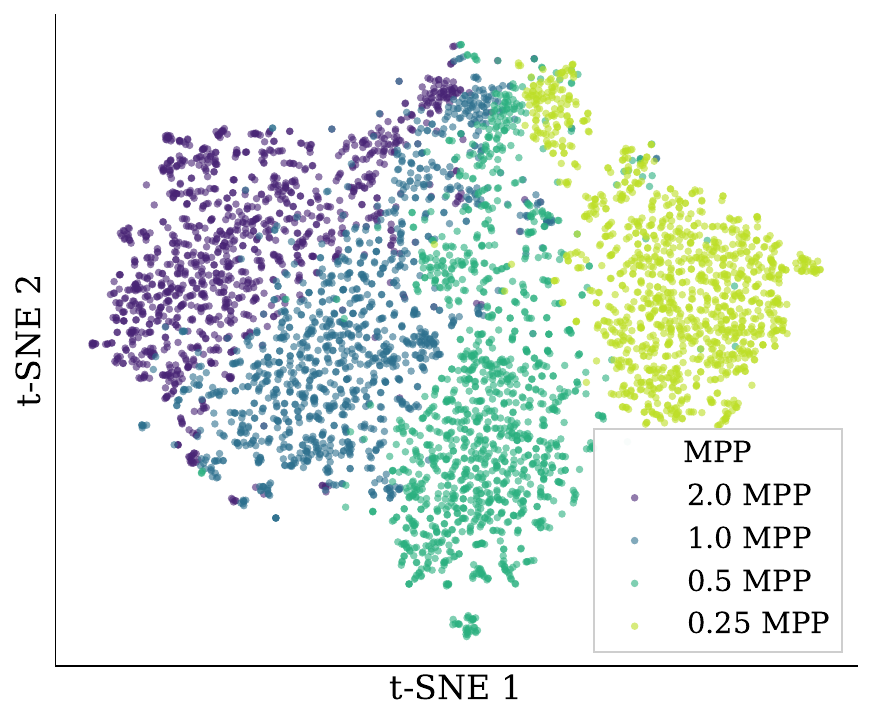}
\end{subfigure}
\caption{MMPP (DinoV2 AUG) INFOMAX10}
\end{figure}

\begin{figure}[!ht]
\centering
\begin{subfigure}[c]{0.35\textwidth}
    \includegraphics[width=\textwidth]{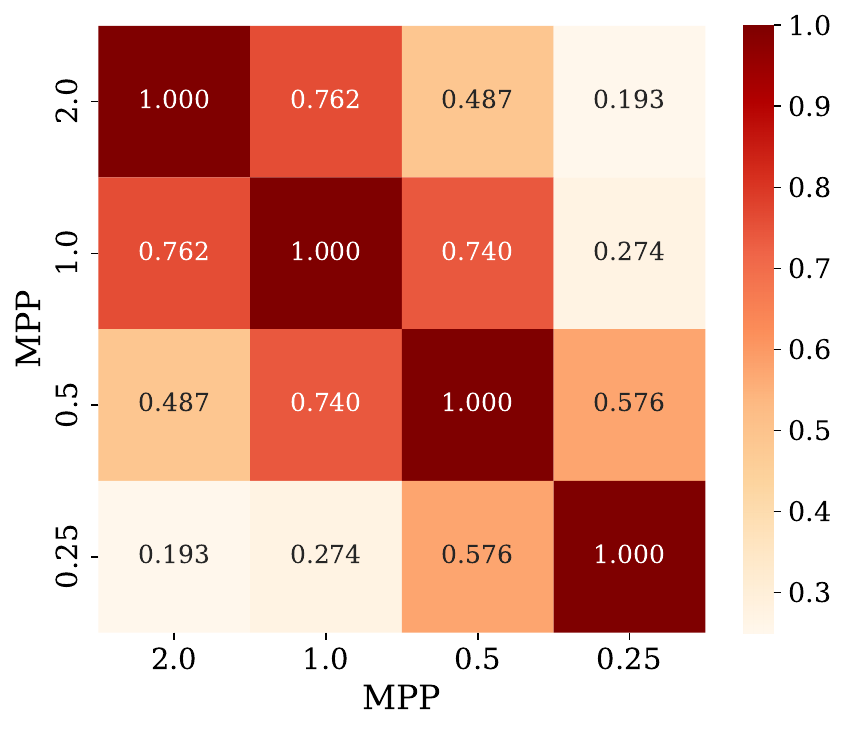}
\end{subfigure}
\hfill
\begin{subfigure}[c]{0.35\textwidth}
    \includegraphics[width=\textwidth]{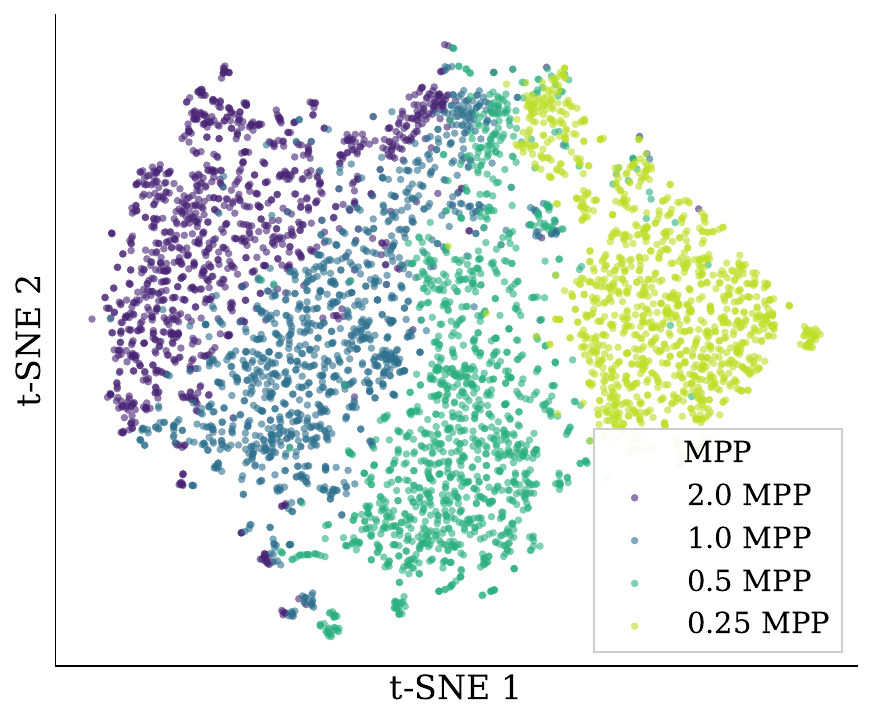}
\end{subfigure}
\caption{MMPP (DinoV2 AUG) CU}
\end{figure}

\clearpage

\begin{figure}[!ht]
\centering
\begin{subfigure}[c]{0.35\textwidth}
    \includegraphics[width=\textwidth]{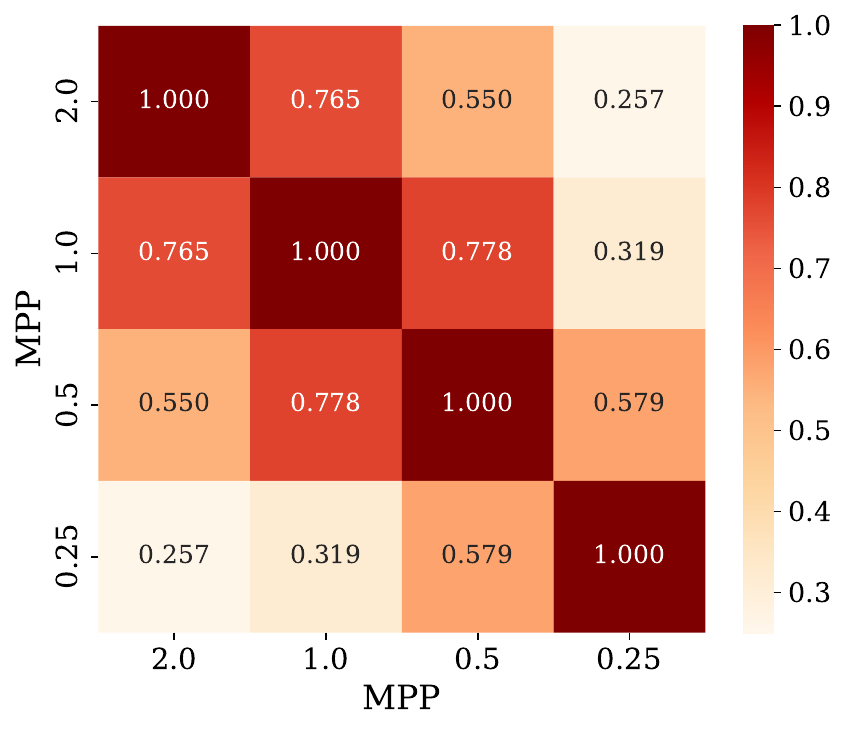}
\end{subfigure}
\hfill
\begin{subfigure}[c]{0.35\textwidth}
    \includegraphics[width=\textwidth]{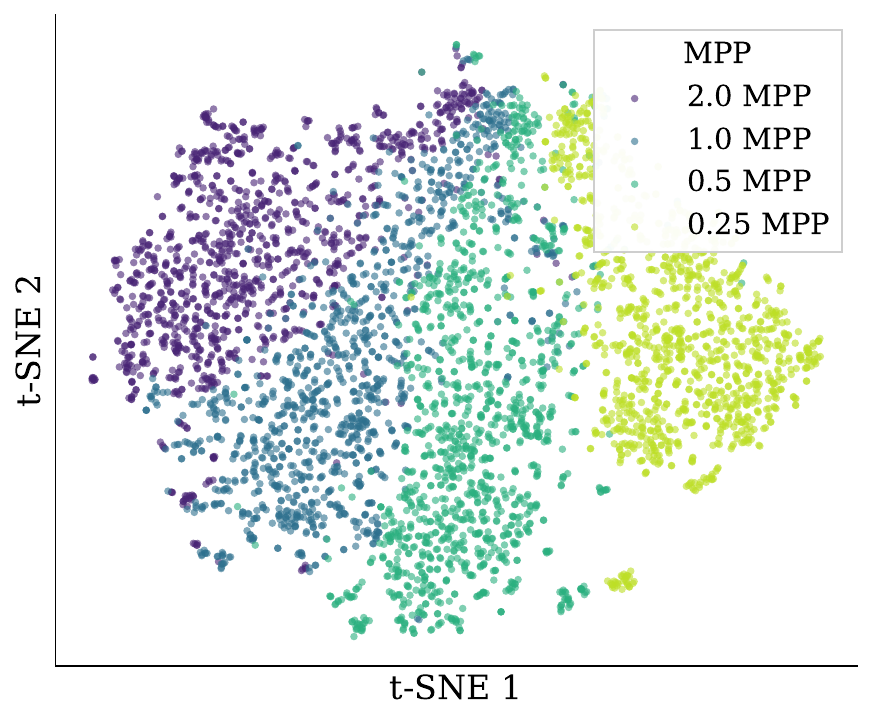}
\end{subfigure}
\caption{MMPP (DinoV2 AUG) INFO\_MINMAX}
\end{figure}

\begin{figure}[!ht]
\centering
\begin{subfigure}[c]{0.35\textwidth}
    \includegraphics[width=\textwidth]{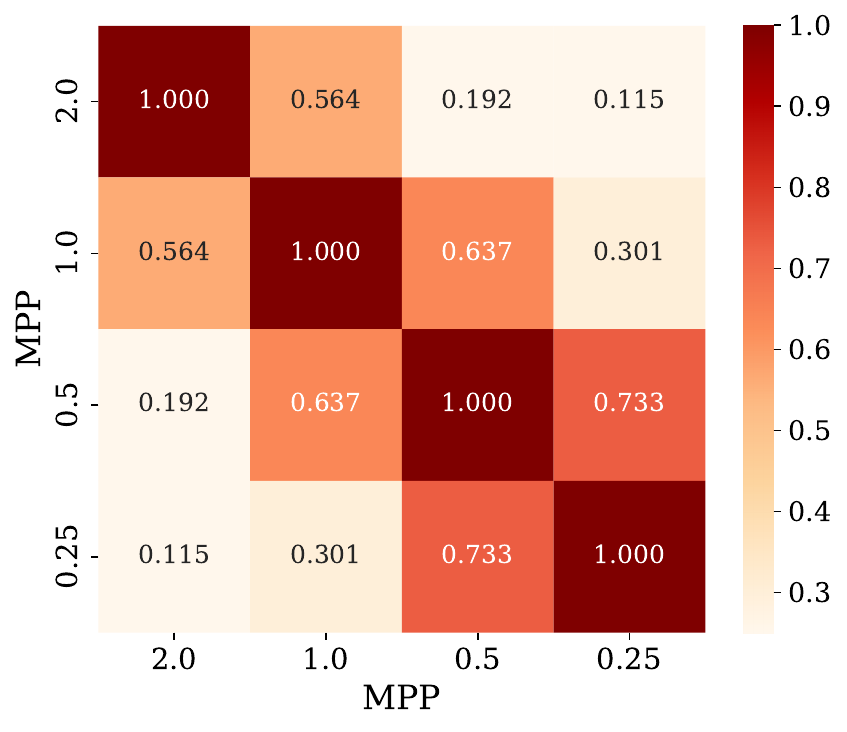}
\end{subfigure}
\hfill
\begin{subfigure}[c]{0.35\textwidth}
    \includegraphics[width=\textwidth]{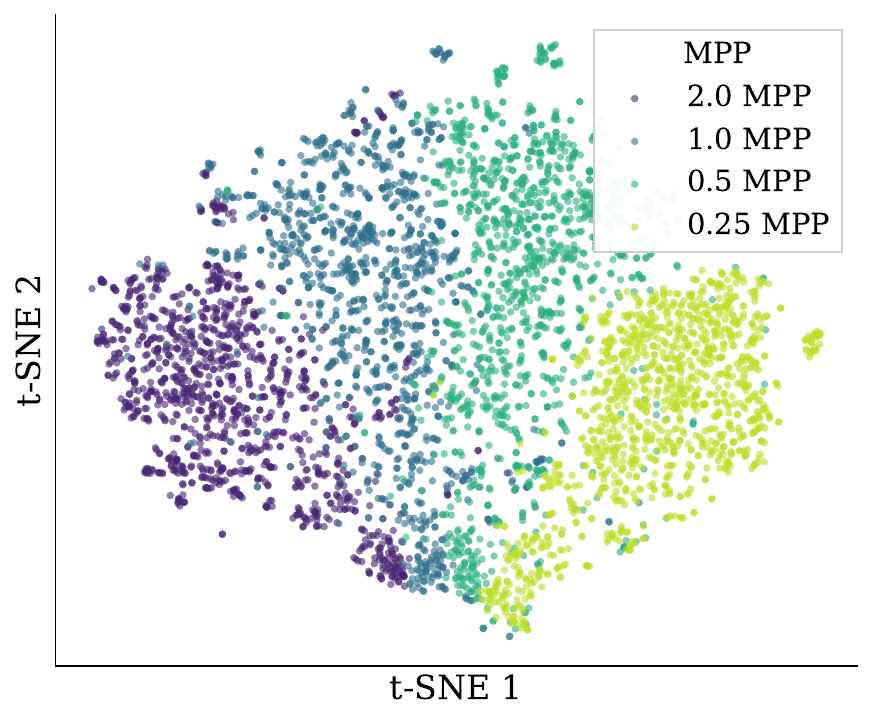}
\end{subfigure}
\caption{1 MPP (DinoV2 AUG)}
\end{figure}

\begin{figure}[!ht]
\centering
\begin{subfigure}[c]{0.35\textwidth}
    \includegraphics[width=\textwidth]{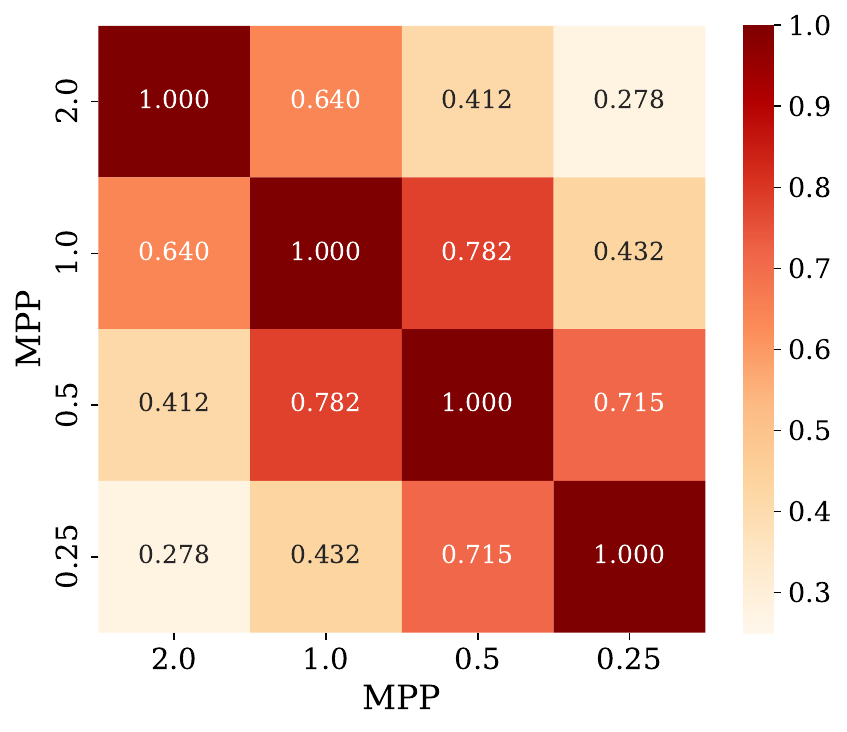}
\end{subfigure}
\hfill
\begin{subfigure}[c]{0.35\textwidth}
    \includegraphics[width=\textwidth]{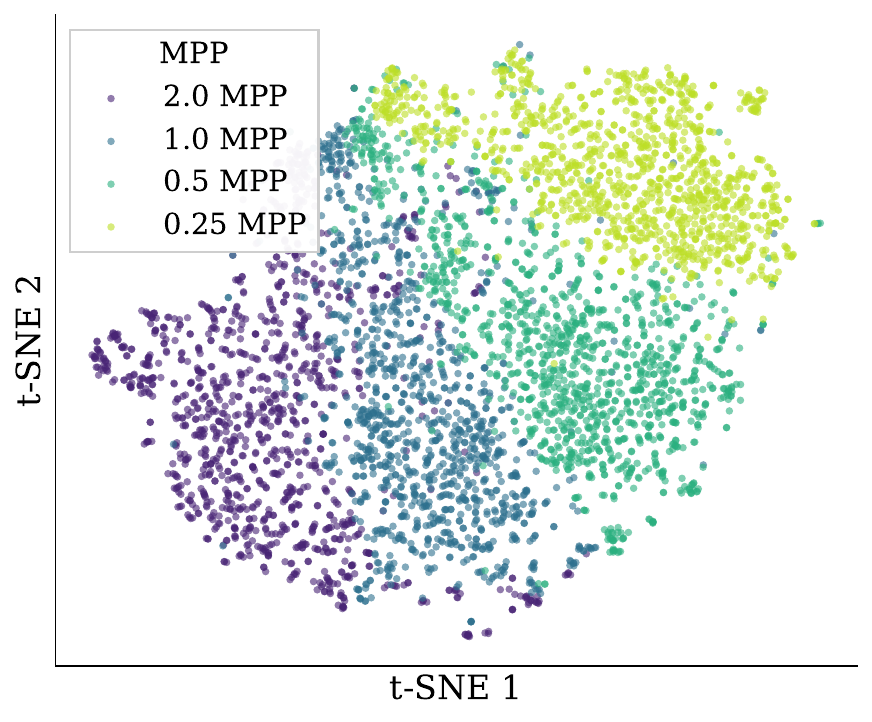}
\end{subfigure}
\caption{2 MPP (DinoV2 AUG)}
\end{figure}

\clearpage

\begin{figure}[!ht]
\centering
\begin{subfigure}[c]{0.35\textwidth}
    \includegraphics[width=\textwidth]{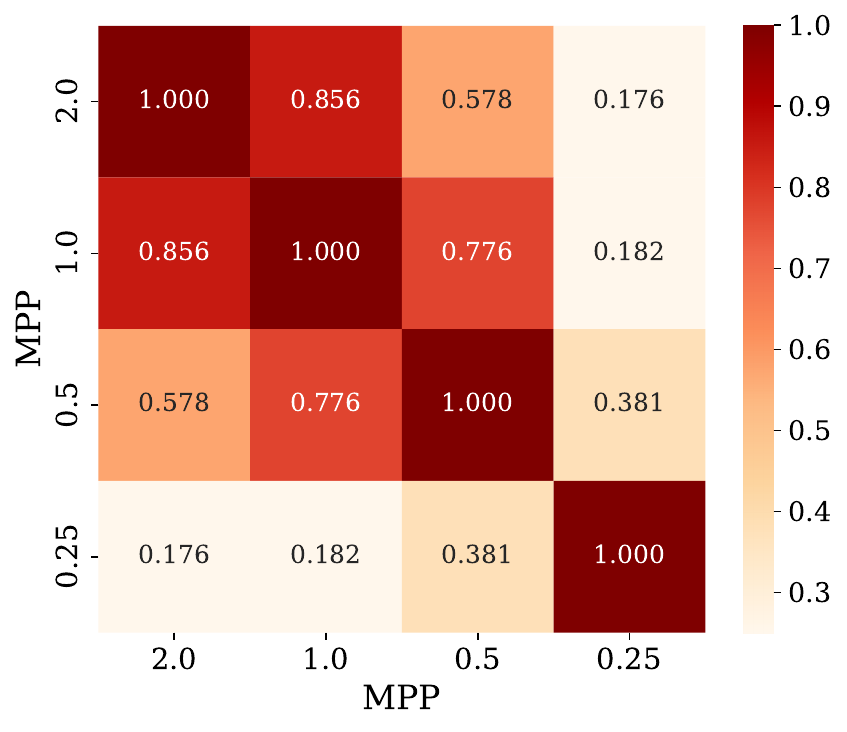}
\end{subfigure}
\hfill
\begin{subfigure}[c]{0.35\textwidth}
    \includegraphics[width=\textwidth]{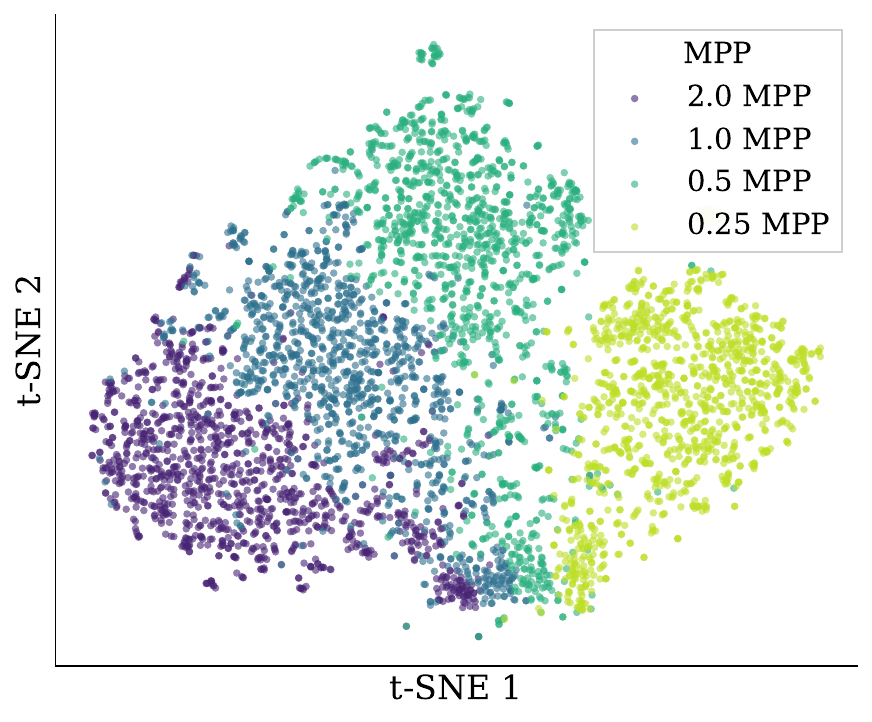}
\end{subfigure}
\caption{0.25 MPP (DinoV2 AUG)}
\end{figure}

\section{Downstream Performance Controlled Experiments}  \label{apd:downstream_results_vits}

\begin{table}[h!]
\centering
\begin{tabular}{lcccc}
\hline
\textbf{MPP} & \textbf{DU} & \textbf{CU} & \textbf{CU-MAXAVG} & \textbf{CU-MINMAX} \\
\hline
0.25 & 31.14$\pm$0.58 & 30.83$\pm$0.56 & 31.06$\pm$0.44 & 30.91$\pm$0.41 \\
0.375 & 32.75$\pm$0.31 & 33.68$\pm$0.25 & 33.07$\pm$0.44 & 33.53$\pm$0.17 \\
0.5 & 34.84$\pm$0.27 & 36.22$\pm$0.38 & 35.82$\pm$0.35 & 35.27$\pm$0.27 \\
0.75 & 37.18$\pm$0.49 & 38.61$\pm$0.40 & 38.57$\pm$0.44 & 37.73$\pm$0.34 \\
1.0 & 41.49$\pm$0.40 & 42.47$\pm$0.50 & 41.90$\pm$0.47 & 41.16$\pm$0.26 \\
1.5 & 38.35$\pm$0.53 & 42.21$\pm$0.74 & 41.45$\pm$0.72 & 40.44$\pm$0.48 \\
2.0 & 46.22$\pm$0.68 & 47.28$\pm$0.66 & 46.75$\pm$0.89 & 46.12$\pm$0.57 \\
\hline
\end{tabular}
\caption{KNN Balanced Accuracy for BRACS by MPP Value (mean $\pm$ standard error).}
\end{table}

\begin{table}[h!]
\centering
\begin{tabular}{lcccc}
\hline
\textbf{MPP} & \textbf{DU} & \textbf{CU} & \textbf{CU-MAXAVG} & \textbf{CU-MINMAX} \\
\hline
0.25 & 33.12$\pm$0.34 & 34.14$\pm$0.61 & 33.71$\pm$0.63 & 33.03$\pm$0.48 \\
0.375 & 37.13$\pm$0.68 & 38.80$\pm$0.61 & 38.21$\pm$0.49 & 37.46$\pm$0.36 \\
0.5 & 39.41$\pm$0.79 & 40.51$\pm$0.45 & 40.41$\pm$0.32 & 40.39$\pm$0.29 \\
0.75 & 44.04$\pm$0.47 & 44.85$\pm$0.57 & 45.73$\pm$0.79 & 43.96$\pm$0.43 \\
1.0 & 45.87$\pm$0.80 & 46.57$\pm$0.54 & 47.74$\pm$0.82 & 46.61$\pm$0.54 \\
1.5 & 47.88$\pm$1.10 & 48.08$\pm$0.69 & 49.90$\pm$1.11 & 48.00$\pm$0.58 \\
2.0 & 46.04$\pm$1.03 & 47.95$\pm$0.46 & 48.88$\pm$1.00 & 46.45$\pm$0.59 \\
\hline
\end{tabular}
\caption{LOGREG Balanced Accuracy for BRACS by MPP Value (mean $\pm$ standard error).}
\end{table}
\begin{table}[h!]
\centering
\begin{tabular}{lcccc}
\hline
\textbf{MPP} & \textbf{DU} & \textbf{CU} & \textbf{CU-MAXAVG} & \textbf{CU-MINMAX} \\
\hline
0.25 & 20.44$\pm$0.18 & 20.72$\pm$0.05 & 20.58$\pm$0.08 & 20.72$\pm$0.39 \\
0.375 & 20.03$\pm$0.78 & 22.21$\pm$0.20 & 22.35$\pm$0.15 & 22.20$\pm$0.13 \\
0.5 & 25.08$\pm$0.46 & 25.21$\pm$0.13 & 25.72$\pm$0.21 & 25.07$\pm$0.20 \\
0.75 & 24.82$\pm$0.34 & 26.24$\pm$0.15 & 25.78$\pm$0.39 & 25.19$\pm$0.18 \\
1.0 & 26.84$\pm$0.58 & 27.46$\pm$0.18 & 27.73$\pm$0.20 & 26.79$\pm$0.34 \\
1.5 & 26.10$\pm$0.61 & 28.38$\pm$0.34 & 28.80$\pm$0.38 & 27.24$\pm$0.32 \\
2.0 & 27.21$\pm$0.27 & 28.46$\pm$0.19 & 28.65$\pm$0.48 & 27.69$\pm$0.09 \\
\hline
\end{tabular}
\caption{KNN Balanced Accuracy for TCGA by MPP Value (mean $\pm$ standard error).}
\end{table}
\begin{table}[h!]
\centering
\begin{tabular}{lcccc}
\hline
\textbf{MPP} & \textbf{DU} & \textbf{CU} & \textbf{CU-MAXAVG} & \textbf{CU-MINMAX} \\
\hline
0.25 & 19.94$\pm$0.11 & 20.34$\pm$0.16 & 20.10$\pm$0.24 & 20.29$\pm$0.11 \\
0.375 & 19.23$\pm$0.55 & 21.28$\pm$0.34 & 21.89$\pm$0.45 & 20.36$\pm$0.32 \\
0.5 & 23.13$\pm$0.50 & 24.07$\pm$0.34 & 24.30$\pm$0.15 & 23.21$\pm$0.37 \\
0.75 & 22.44$\pm$0.56 & 25.93$\pm$0.40 & 26.14$\pm$0.20 & 23.95$\pm$1.02 \\
1.0 & 25.78$\pm$0.73 & 26.82$\pm$0.52 & 26.86$\pm$0.07 & 26.01$\pm$0.40 \\
1.5 & 22.91$\pm$0.40 & 26.49$\pm$0.08 & 27.28$\pm$0.24 & 24.74$\pm$0.86 \\
2.0 & 24.67$\pm$0.22 & 26.18$\pm$0.52 & 27.33$\pm$0.17 & 25.23$\pm$0.64 \\
\hline
\end{tabular}
\caption{LOGREG Balanced Accuracy for TCGA by MPP Value (mean $\pm$ standard error).}
\end{table}

\end{document}